\documentclass[runningheads]{llncs}
\usepackage{graphicx}
\usepackage{amsmath,amssymb}
\usepackage{color}

\usepackage{comment}
\usepackage[width=122mm,left=12mm,paperwidth=146mm,height=193mm,top=12mm,paperheight=217mm]{geometry}

\usepackage{color}
\definecolor{green}{rgb}{0, 0.5, 0}
\definecolor{orange}{rgb}{0.8, 0.6, 0.2}
\definecolor{red}{rgb}{1.0, 0.0, 0.0}
\definecolor{teal}{rgb}{0.0, 0.4, 0.4}
\definecolor{purple}{rgb}{0.65,0,0.65}
\definecolor{saffron}{rgb}{0.95,0.75,0.2}
\definecolor{turquoise}{rgb}{0.0,0.5,0.5}

\usepackage[ruled,vlined,linesnumbered]{algorithm2e}
\usepackage{multirow}

\newcommand{\kev}[1]{{\color{red}#1}}
\newcommand{\tom}[1]{{\color{green}#1}}

\newcommand{\MATTHIAS}[1]{\textcolor{red}{\textbf{\{Matthias: #1\}}}}

\newcommand{\blue}[1]{{\color{blue}\textbf{#1}}}
\newcommand{\green}[1]{{\color{green}\textbf{#1}}}

\usepackage[normalem]{ulem}
\newcommand{\scratch}[1]{{\color{red}\sout{#1}}}

\usepackage{mathtools}
\usepackage{amsmath, amssymb, amsfonts, amsopn}
\usepackage{graphicx} 
\usepackage{textcomp}
\usepackage{xfrac}
\usepackage{bbm}

\usepackage{overpic}
\usepackage{subfig}
\usepackage{wrapfig}

\newcommand{\textapprx}{\raisebox{0.1ex}{\texttildelow}}
\newcommand{\etal}{et al.}
\newcommand{\hidecomment}[1]{}

\newcommand{\num}{{\color{blue}[X]~}}

\begin{document}

\title{PlaneMatch: Patch Coplanarity Prediction for Robust RGB-D Reconstruction} 
\titlerunning{PlaneMatch}

\author{Yifei Shi\inst{1,2} \and
Kai Xu \inst{1,2}\thanks{Corresponding author: kevin.kai.xu@gmail.com} \and
Matthias Nie{\ss}ner\inst{3} \and
Szymon Rusinkiewicz\inst{1} \and
Thomas Funkhouser\inst{1,4}}
\authorrunning{Yifei Shi et al.}

\institute{Princeton University\and
National University of Defense Technology\and
Technical University of Munich\and
Google}

\maketitle


\begin{abstract}
We introduce a novel RGB-D patch descriptor designed for detecting coplanar surfaces in SLAM reconstruction.
The core of our method is a deep convolutional neural network that takes in RGB, depth, and normal information of a planar patch in an image and outputs a descriptor that can be used to find coplanar patches from other images.
We train the network on 10 million triplets of coplanar and non-coplanar patches, and evaluate on
a new coplanarity benchmark created from commodity RGB-D scans.  Experiments show that our learned
descriptor outperforms alternatives extended for this new task by a significant margin.
In addition, we demonstrate the benefits of coplanarity matching in a robust RGBD reconstruction formulation.
We find that coplanarity constraints detected with our method are sufficient to get reconstruction results
comparable to state-of-the-art frameworks on most scenes, but outperform other methods on established benchmarks when combined with traditional keypoint matching.

\end{abstract} 


\section{Introduction}

With the recent proliferation of inexpensive RGB-D sensors, it is now becoming practical for people to scan 3D models of large indoor environments with hand-held cameras, enabling applications in cultural heritage, real estate, virtual reality, and many other fields.
Most state-of-the-art RGB-D reconstruction algorithms either perform frame-to-model alignment \cite{izadi2011kinectfusion} or match keypoints for global pose estimation \cite{dai2017bundlefusion}.
Despite the recent progress in these algorithms, registration of hand-held RGB-D scans remains challenging when local surface features are not discriminating and/or when scanning loops have little or no overlaps.


An alternative is to detect planar features and associate them across frames with coplanarity, parallelism, and perpendicularity constraints \cite{zhang2015online,halber2016fine,lee2017joint,ma2016cpa,trevor2012planar,zhang2016emptying,huang2017dlight}.
Recent work has shown compelling evidence that planar patches can be detected and tracked robustly, especially in indoor environments where flat surfaces are ubiquitous.
In cases where traditional features such as keypoints are missing (e.g., wall), there seems tremendous potential to support existing 3D reconstruction pipelines.

Even though coplanarity matching is a promising direction, current approaches lack strong per-plane feature descriptors for establishing putative matches between disparate observations.
As a consequence, coplanarity priors have only been used in the context of frame-to-frame tracking \cite{zhang2015online} or in post-process steps for refining a global optimization \cite{halber2016fine}.
We see this as analogous to the relationship between ICP and keypoint matching:
just as ICP only converges with a good initial guess for pose, current methods for exploiting coplanarity are unable to initialize a reconstruction process from scratch due to the lack of discriminative coplanarity features.


\begin{figure}[t!] \centering
    \begin{overpic}[width=0.8\linewidth]{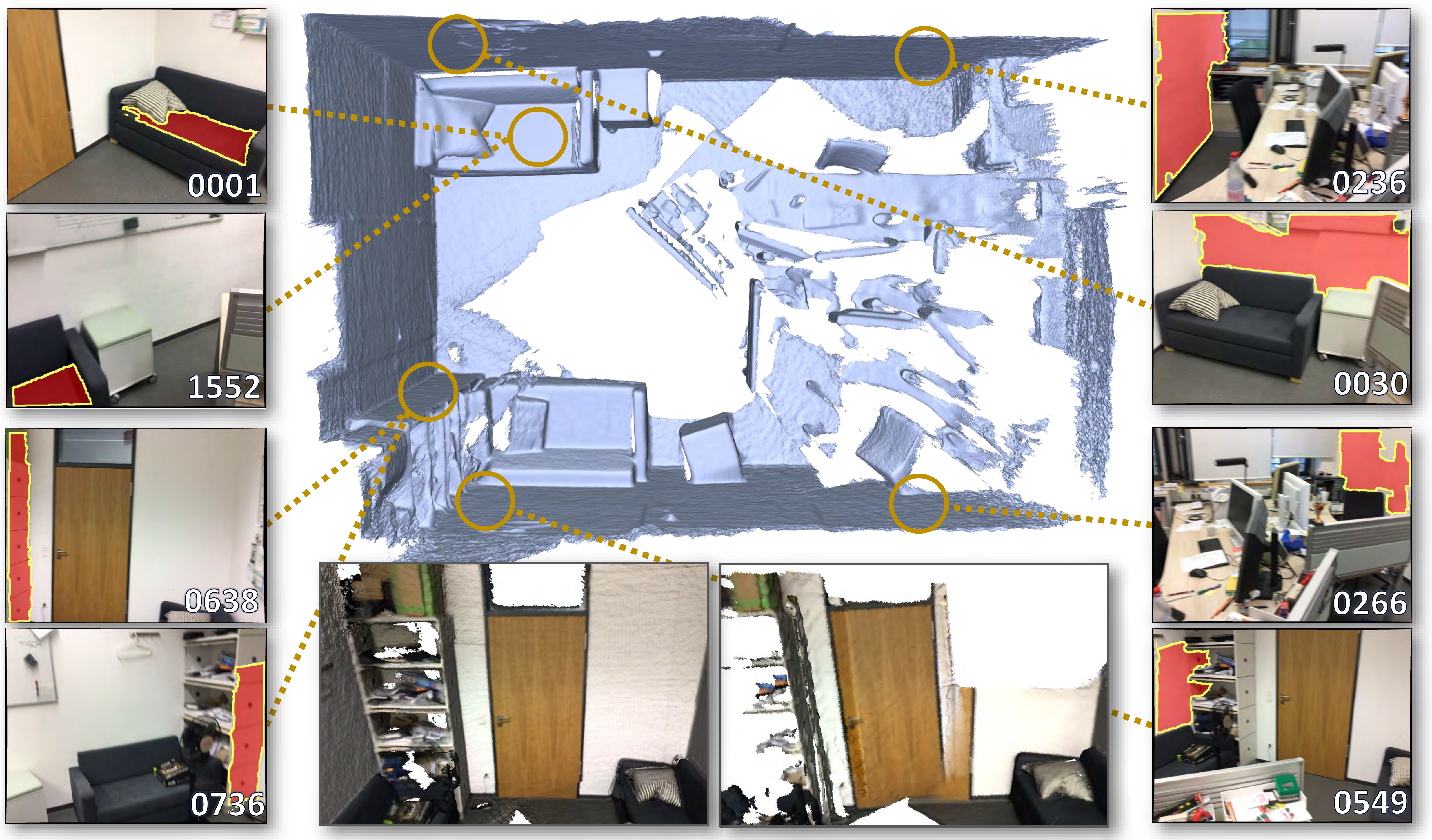}
    \end{overpic}
    \caption{Scene reconstruction based on coplanarity matching of patches across different views (numbers indicate frame ID) for both overlapping (left two pairs) and non-overlapping (right two pairs) patch pairs.
    The two pairs to the right are long-range, without overlapping.
    The bottom shows a zoomed-in comparison between our method (left) and key-point matching based method~\protect~\cite{dai2017bundlefusion} (right).}
    \label{fig:teaser}
\end{figure} 

This paper aims to enable global, \emph{ab initio} coplanarity matching by introducing a discriminative feature descriptor for planar patches of RGB-D images.
Our descriptor is learned from data to produce features whose L2 difference is predictive of \emph{whether or not two RGB-D patches from different frames are coplanar}.
It can be used to detect pairs of coplanar patches in RGB-D scans {\em without an initial alignment}, which can be used to find loop closures or to provide coplanarity constraints for global alignment (see Figure~\ref{fig:teaser}).

A key novel aspect of this approach is that it focuses on detection of coplanarity rather than overlap.
As a result, our plane patch features can be used to discover long-range alignment constraints (like ``loop closures'') between distant, non-overlapping parts of the same large surface (e.g., by recognizing carpets on floors, tiles on ceilings, paneling on walls, etc.).
In Figure~\ref{fig:teaser}, the two patch pairs shown to the right helped
produce a reconstruction with globally flat walls.

To learn our planar patch descriptor, we design a deep network that takes in color, depth, normals, and multi-scale context for pairs of planar patches extracted from RGB-D images, and predicts whether they are coplanar or not.
The network is trained in a self-supervised fashion where training examples are automatically extracted from coplanar and noncoplanar patches from ScanNet~\cite{dai2017scannet}.

In order to evaluate our descriptor, we introduce a new coplanarity matching datasets, where we can see in series of thorough experiments that our new descriptor outperforms existing baseline alternatives by significant margins.
Furthermore, we demonstrate that by using our new descriptor, we are able to compute strong coplanarity constraints that improve the performance of current global RGB-D registration algorithms.
In particular, we show that by combining coplanarity and point-based correspondences reconstruction algorithms are able to handle difficult cases, such as scenes with a low number of features or limited loop closures.
We outperform other state-of-the-art algorithms on the standard TUM RGB-D reconstruction benchmark \cite{sturm12iros}.
Overall, the research contributions of this paper are:
\vspace{-0.2cm}
\begin{itemize}
\setlength{\topsep}{0pt}
\setlength{\parsep}{0pt}
\setlength{\parskip}{0pt}
\setlength{\itemsep}{2pt}

\item A new task: predicting coplanarity of image patches for the purpose of RGB-D image registration.

\item A self-supervised process for training a deep network to produce
features for predicting whether two image patches are coplanar or not.

\item An extension of the robust optimization algorithm \cite{choi2015robust} to solve camera poses with coplanarity constraints. 

\item A new training and test benchmark for coplanarity prediction.

\item Reconstruction results demonstrating that coplanarity can be used to align scans where keypoint-based methods fail to find loop closures.

\end{itemize}

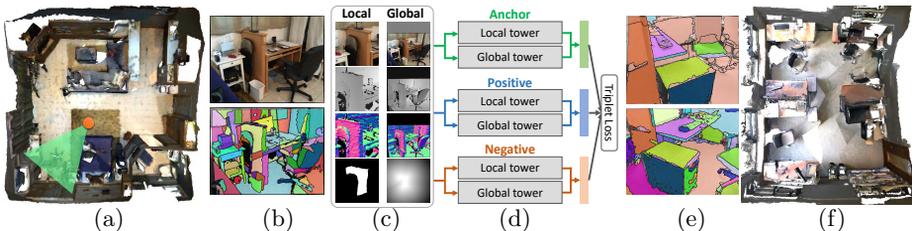
\begin{figure*}[t!] \centering
    \begin{overpic}[width=0.999\linewidth]{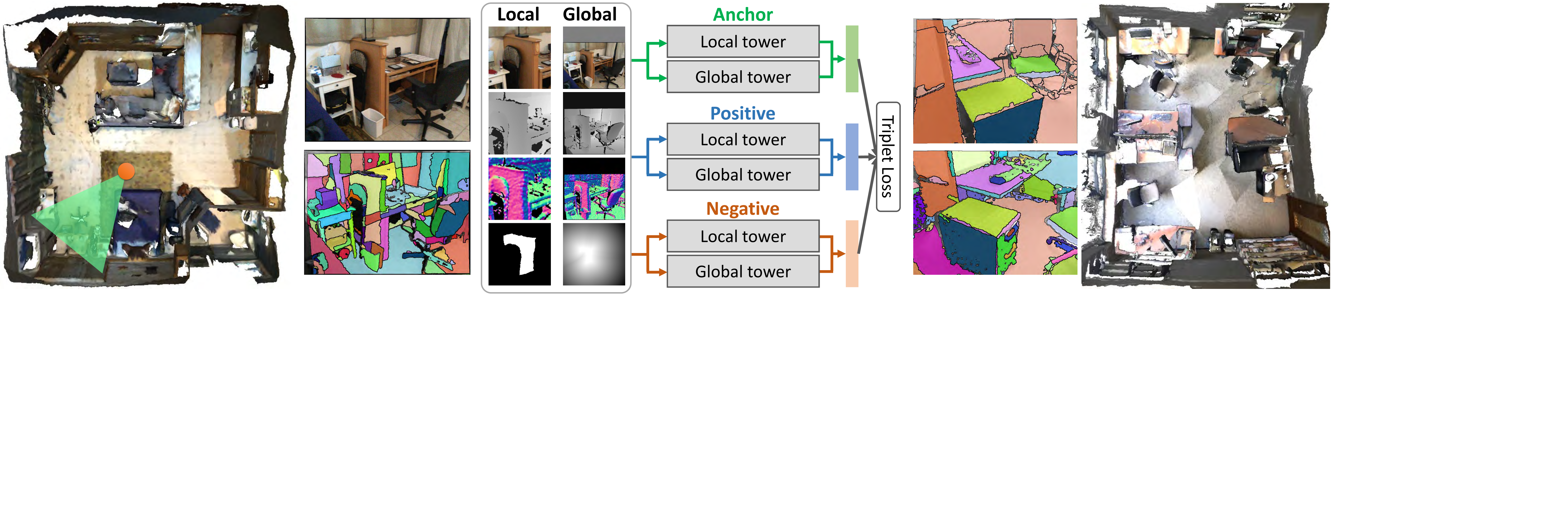}
    \small
        \put(10,-2){(a)}
        \put(28.5,-2){(b)}
        \put(40.5,-2){(c)}
        \put(54.5,-2){(d)}
        \put(74,-2){(e)}
        \put(89.5,-2){(f)}
    \end{overpic}
    \caption{An overview of our method. We train an embedding network (c-d)
to predict coplanarity for a pair of planar patches across different views, based on the co-planar patches (b)
sampled from training sequences with ground-truth camera poses (a).
Given a test sequence, our robust optimization performs reconstruction (f) based on predicted co-planar patches (e).}
    \label{fig:overview}
\end{figure*} 

\section{Related Work}

\vspace*{1mm}\noindent{\bf RGB-D Reconstruction:} Many SLAM systems have
been described for reconstructing 3D scenes from RGB-D video.
Examples include KinectFusion
\cite{Newcombe2011_KinfuMapTrack,izadi2011kinectfusion}, VoxelHashing
\cite{niessner2013hashing}, ScalableFusion \cite{chen2013scalable},
Point-based Fusion \cite{Keller3DV2013}, Octrees on CPU
\cite{Steinbruecker-etal-icra14}, Elastic Fusion \cite{Whelan15rss},
Stereo DSO \cite{wang2017stereo}, 
Colored Registration \cite{park2017colored},
and Bundle Fusion
\cite{dai2017bundlefusion}.
These systems generally perform well for scans with many loop closures and/or when robust IMU
measurements are available.
However, they often exhibit drift in long
scans when few constraints can be established between disparate
viewpoints.
In this work, we detect and enforce coplanarity constraints between planar patches to address this issue as an alternative feature channel for global matching.

\vspace*{1mm}\noindent{\bf Feature Descriptors:} Traditionally, SLAM systems
have utlized {\em keypoint} detectors and descriptors to establish
correspondence constraints for camera pose estimation.  Example
keypoint descriptors include SIFT \cite{lowe1999object}, SURF
\cite{bay2006surf}, ORB \cite{rublee2011orb}, etc.  More recently,
researchers have learned keypoint descriptors from data -- e.g.,
MatchNet \cite{han2015matchnet}, Lift \cite{yi2016lift}, SE3-Nets
\cite{byravan2017se3}, 3DMatch
\cite{zeng20173dmatch}, Schmidt et al.~\cite{schmidt2017self}.  These methods rely upon repeatable extraction
of keypoint positions, which is difficult for widely disparate views.
In contrast, we explore the more robust method of extracting planar patches
without concern for precisely positioning the patch center.

\vspace*{1mm}\noindent{\bf Planar Features:} Many previous papers have leveraged
planar surfaces for RGB-D reconstruction.
The most commmon approach
is to detect planes in RGB-D scans, establish
correspondences between matching features, and solve for
the camera poses that align the corresponding features \cite{concha2015dpptam,dou2012exploring,hsiao2017keyframe,pietzsch2008planar,proencca2017probabilistic,salas2014dense,taguchi2013point,weingarten20063d}.
More recent approaches build models comprising planar patches, possibly
with geometric constraints \cite{halber2016fine,stuckler2008orthogonal}, and
match planar features found in scans to planar patches in the models
\cite{halber2016fine,lee2017joint,ma2016cpa,trevor2012planar,zhang2016emptying}.
The search for correspondences is often aided by hand-tuned
descriptors designed to detect overlapping surface regions. 
In contrast, our approach finds correspondences between
\emph{coplanar patches} (that may not be overlapping); we learn descriptors for this task with a deep network.

\vspace*{1mm}\noindent{\bf Global Optimization:} For large-scale surface reconstruction,
it is common to use off-line or asynchronously executed
global registration procedures.  A common formulation is to compute a pose graph with edges representing pairwise transformations between frames and then optimize an objective function penalizing deviations from these pairwise alignments \cite{grisetti2010tutorial,henry2010rgb,zhou2013dense}.  Recent methods \cite{choi2015robust,zhou2016fast} use indicator variables to identify loop closures or matching points during global optimization using a least-squares formulation. 
We extend this formulation by setting indicator variables for individual coplanarity constraints.


\section{Method}
Our method consists of two components: 1) a deep neural network trained to generate a descriptor that can be used to discover coplanar pairs of RGB-D patches without an initial registration, and 2) a global SLAM reconstruction algorithm that takes special advantage of detected pairs of coplanar patches.

\subsection{Coplanarity Network}

Coplanarity of two planar patches is by definition geometrically measurable. However, for two patches
that are observed from different, yet unknown views, whether they are coplanar is not determinable based on geometry alone.
Furthermore, it is not clear that coplanarity can be deduced solely from the local appearance of the imaged objects.
We argue that the prediction of coplanarity across different views is a structural, or even semantic,
visual reasoning task, for which neither geometry nor local appearance alone is reliable.

Humans infer coplanarity by perceiving and understanding the structure
and semantics of objects and scenes, and contextual information plays a critical role in this reasoning task.
For example, humans are able to differentiate different facets of an object, from virtually \emph{any} view,
by reasoning about the structure of the facets and/or by relating them to surrounding objects.
Both involve inference with a context around the patches being considered, possibly at multiple scales.
This motivates us to learn to predict cross-view coplanarity from appearance and geometry, using
\emph{multi-scale contextual information}.
We approach this task by learning an embedding network that maps coplanar patches
from different views nearby in feature space.

\begin{figure}[t!] \centering
    \begin{overpic}[width=0.85\linewidth]{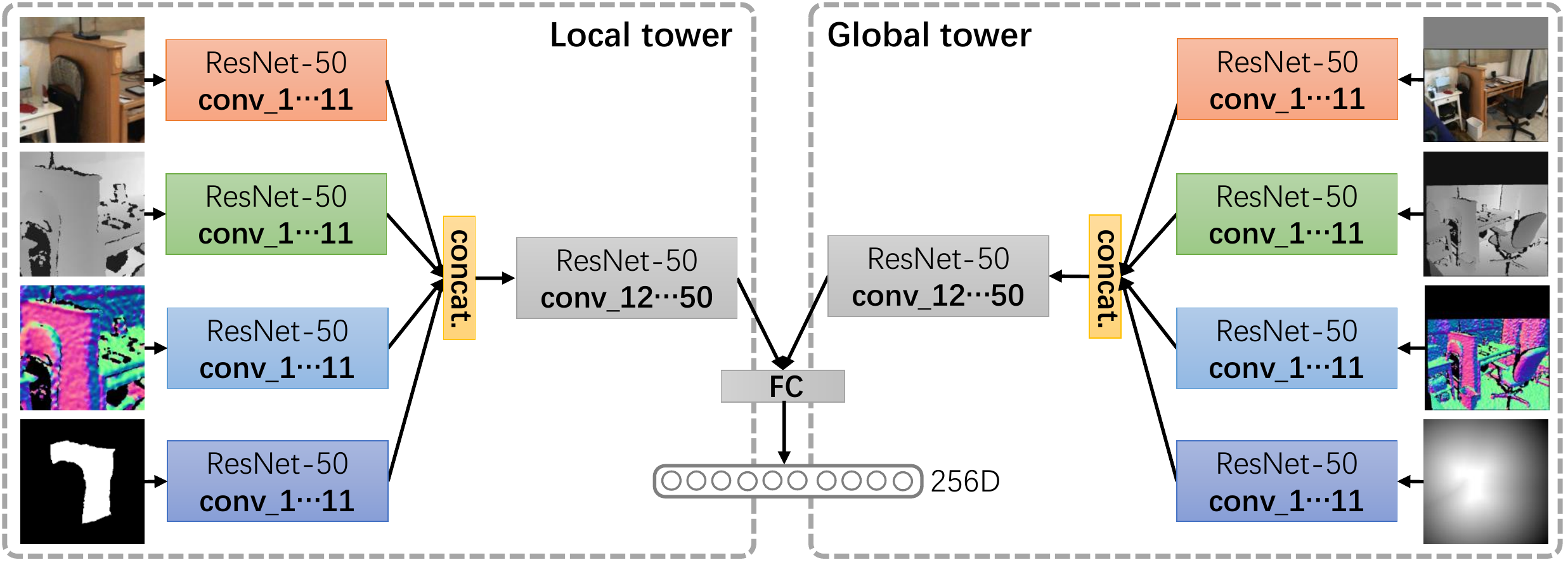}
    \end{overpic}
    \caption{Network architecture of the local and global towers.
    Layers shaded in the same color share weights.}
    \label{fig:network}
\end{figure}

\vspace*{2mm}\noindent{\bf Network Design:}
Our coplanarity network (Figure~\ref{fig:overview} and~\ref{fig:network})
is trained with triplets of planar patches,
each involving an anchor, a coplanar patch (positive) and a noncoplanar patch (negative), similar to~\cite{schroff2015facenet}.
Each patch of a triplet is fed into a convolutional network based on ResNet-50~\cite{he2016deep}
for feature extraction, and a triplet loss is estimated based on the relative proximities of the three features.
To learn coplanarity from both appearance and geometry,
our network takes multiple channels as input: an RGB image, depth image, and normal map.

We encode the contextual information of a patch at two scales,
local and global.
This is achieved by cropping the input images (in all channels)
to rectangles of $1.5$ and $5$ times the size
of the patch's bounding box, respectively.
All cropped images are clamped at image boundaries,
padded to a square, and then resized to $224\times 224$.
The padding uses 50\% gray for RGB images and a value of $0$ for depth and normal maps;
see Figure~\ref{fig:network}.

To make the network aware of the region of interest (as opposed to context) in each input image,
we add, for each of the two scales, an extra binary mask channel.
The local mask is binary, with the patch of interest in white and the rest of the image in black.
The global mask, in contrast, is continuous, with the patch of interest in white and then
a smooth decay to black outside the patch boundary.
Intuitively, the local mask helps the network distinguish the patch of interest from the close-by
neighborhood, e.g.~other sides on the same object.
The global mask, on the other hand, directs the network to learn global structure by
attending to a larger context, with importance smoothly decreasing
based on distance to the patch region.
Meanwhile, it also weakens the effect of specific patch shape, which is unimportant
when considering global structure.

In summary, each scale consists of RGB, depth, normal, and mask channels.
These inputs are first encoded independently.
Their feature maps are concatenated after the $11$-th convolutional layer,
and then pass through the remaining $39$ layers.
The local and global scales share weights for the corresponding channels,
and their outputs are finally combined with a fully connected layer (Figure~\ref{fig:network}).

\vspace*{2mm}\noindent{\bf Network Training:}
%
The training data for our network are generated from datasets of RGB-D scans of 3D indoor scenes, with high-quality camera poses provided with the datasets.
For each RGB-D frame, we segment it into planar patches using agglomerative
clustering on the depth channel.
For each planar patch, we also estimate its normal based on the depth information.
The extracted patches are projected to image space to generate all the necessary
channels of input to our network.
%
Very small patches, whose local mask image contains less than $300$ pixels with valid depths, are discarded.

\vspace*{2mm}\noindent{\bf Triplet Focal Loss:}
When preparing triplets to train our network, we encounter the well-known problem of a severely imbalanced number of negative and positive patch pairs.
Given a training sequence, there are many more negative pairs, and most of them are too trivial to help the network learn efficiently.
Using randomly sampled triplets would overwhelm the training loss by the easy negatives.

We opt to resolve the imbalance issue by dynamically and discriminatively scaling the losses for hard and easy triplets, inspired by the recent work of focal loss for object detection~\cite{lin2017focal}. Specifically, we propose the \emph{triplet focal loss}:
\begin{equation}\small
L_\text{focal}(x_a,x_p,x_n) = \max \left(0, \frac{\alpha-\Delta d_\text{f}}{\alpha}\right)^\lambda,
\label{eq:focalloss}
\end{equation}
where $x_a$, $x_p$ and $x_n$ are the feature maps extracted for anchor, positive, and negative patches, respectively; $\Delta d_\text{f} = d_\text{f}(x_n, x_a) - d_\text{f}(x_p, x_a)$, with $d_\text{f}$ being the L2 distance between two patch features.  Minimizing this loss encouranges the anchor to be closer to the positive patch than to the negative in descriptor space, but with less weight for larger distances.

\begin{figure}[t!] \centering
    \begin{overpic}[width=0.8\linewidth]{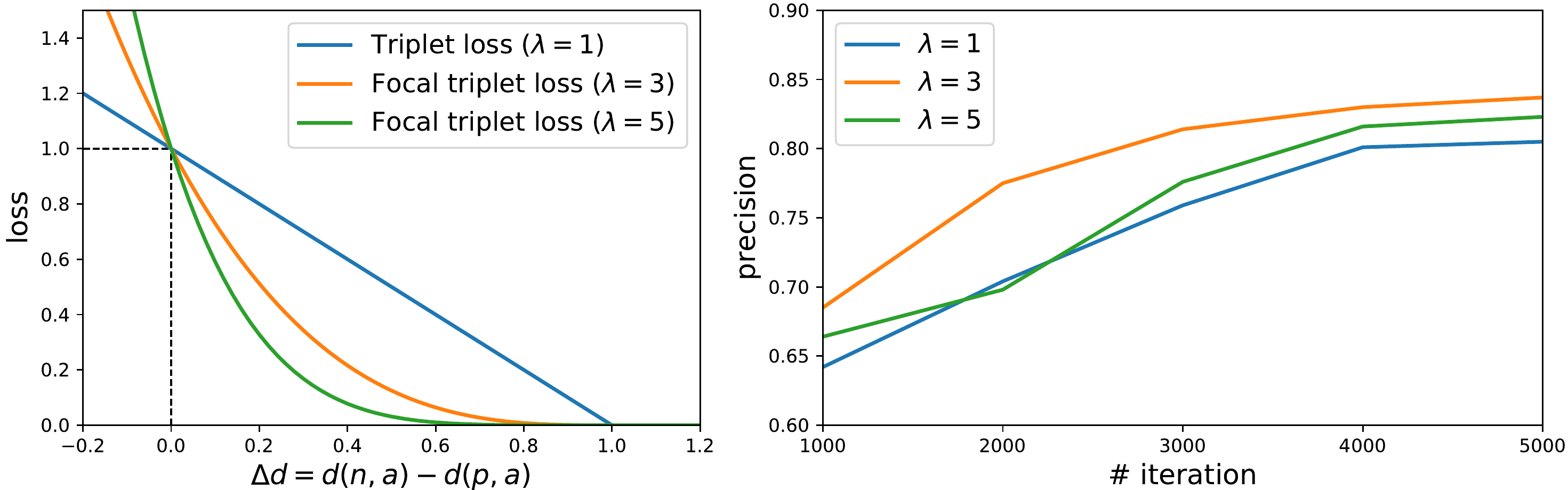}
        \put(19,18){\scalebox{0.77}{$y=\max\left(0,\left(\frac{\alpha-x}{\alpha}\right)\right)^\lambda$}}
        \put(32,14){\scalebox{0.7}{($\alpha=1$)}}
    \end{overpic}
    \caption{Visualization and comparison (prediction accuracy over \#iter.) of different triplet loss functions.}
    \label{fig:focalloss}
\end{figure} 

\begin{figure}[t!] \centering
    \begin{overpic}[width=0.8\linewidth]{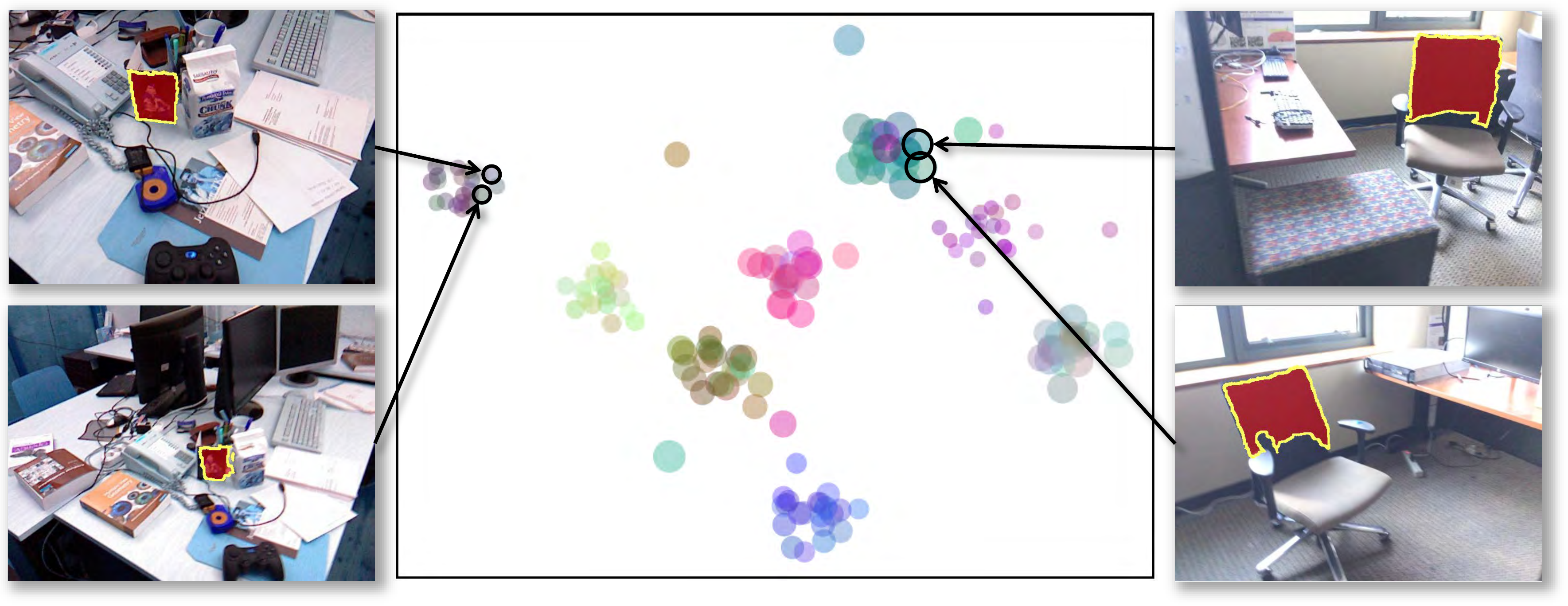}
    \end{overpic}
    \caption{t-SNE visualization of coplanarity-based features of planar patches from different views.
    Ground-truth coplanarity (measured by mutual RMS point-to-plane distance) is encoded by color and physical size of patches by dot size.}
    \label{fig:embedding}
\end{figure}

See Figure~\ref{fig:focalloss}, left, for a visualization of the loss function with $\alpha=1$.
When $\lambda=1$, this loss becomes the usual margined loss,
which gives non-negligible loss to easy examples near the margin $\alpha$. When $\lambda>1$, however, we obtain a focal loss that down-weights easy-to-learn triplets while keeping high loss for hard ones. Moreover, it smoothly adjusts the rate at which easy triplets are down-weighted. We found $\lambda=3$ to achieve the best training efficiency (Figure~\ref{fig:focalloss}, right).
Figure~\ref{fig:embedding} shows a t-SNE visualization of coplanarity-based patch features.

\subsection{Coplanarity-Based Robust Registration}
To investigate the utility of this planar patch descriptor and coplanarity detection
approach for 3D reconstruction, we have developed a global registration algorithm that
estimates camera poses for an RGB-D video using pairwise constraints derived from coplanar
patch matches in addition to keypoint matches.

Our formulation is inspired by the work of Choi~\etal~\cite{choi2015robust}, where
the key feature is the robust penalty term
used for automatically selecting the correct matches from a large pool of hypotheses,
thus avoiding iterative rematching as in ICP.  Note that this formulation does not require an initial alignment of camera poses, which would be required for other SLAM systems that leverage coplanarity constraints.


Given an RGB-D video sequence $\mathcal{F}$,
our goal is to compute for each frame $i \in \mathcal{F}$ a camera pose in the global reference frame, $\mathbf{T}_i=(\mathbf{R}_i,\mathbf{t}_i)$, that brings them into alignment.
This is achieved by jointly aligning each pair of frames $(i,j) \in \mathcal{P}$ that were predicted to have
some set of coplanar patches, $\Pi_{ij}$.
For each pair $\pi=(p, q) \in \Pi_{ij}$,
let us suppose w.l.o.g.~that patch $p$ is from frame $i$ and $q$ from $j$.
Meanwhile, let us suppose a set of matching key-point pairs $\Theta_{ij}$ is detected and matched
between frame $i$ and $j$.
Similarly, we assume for each point pair $\theta=(\mathbf{u}, \mathbf{v}) \in \Theta_{ij}$ that
key-point $\mathbf{u}$ is from frame $i$ and $\mathbf{v}$ from $j$.

\vspace*{2mm}\noindent{\bf Objective Function:}
The objective of our coplanarity-based registration contains four terms, responsible for coplanar alignment, 
coplanar patch pair selection,
key-point alignment,
and key-point pair selection:
\begin{equation}\small
E({T},s) = E_{\text{data-cop}}({T},s)
+ E_{\text{reg-cop}}(s)
+ E_{\text{data-kp}}({T},s)
+ E_{\text{reg-kp}}(s).
\label{eq:robustopt}
\end{equation}

Given a pair of coplanar patches predicted by the network, the \emph{coplanarity data term} enforces the coplanarity, via minimizing the point-to-plane distance from sample points on one patch to the plane defined by the other patch:
\begin{equation}\small
E_{\text{data-cop}}({T},s) = \sum_{(i,j) \in \mathcal{P}}\sum_{\pi \in \Pi_{ij}}{w_\pi\, s_\pi\, \delta^2(\mathbf{T}_i,\mathbf{T}_j,\pi)},
\label{eq:dataterm}
\end{equation}
where $\delta$ is the \emph{coplanarity distance} of a patch pair $\pi=(p,q)$.
It is computed as the root-mean-square point-to-plane distance over both sets of sample points:
$$\small
\delta^2=
\frac{1}{|\mathcal{V}_p|}\,\sum_{\tiny \mathclap {\mathbf{v}_p \in \mathcal{V}_p}}{d^2(\mathbf{T}_i\mathbf{v}_p, \phi^G_q)}
+ \frac{1}{|\mathcal{V}_q|}\,\sum_{\tiny \mathclap{\mathbf{v}_q \in \mathcal{V}_q}}{d^2(\mathbf{T}_j\mathbf{v}_q, \phi^G_p)},
$$
where $\mathcal{V}_p$ is the set of sample points on patch $p$ and
$d$ is point-to-plane distance:
$$\small
d(\mathbf{T}_i\mathbf{v}_p,\phi^G_q) = (\mathbf{R}_i\mathbf{v}_p+\mathbf{t}_i-\mathbf{p}_q) \cdot \mathbf{n}_q.
$$
$\phi^G_q = (\mathbf{p}_q, \mathbf{n}_q)$ is the plane defined by patch $q$, which is
estimated in the \emph{global} reference frame using the corresponding transformation $\mathbf{T}_j$,
and is updated in every iteration.
$s_\pi$~is a control variable (in $[0,1]$) for the selection of patch pair $\pi$, with $1$ standing for selected and $0$ for discarded. $w_\pi$~is a weight that measures the confidence of pair $\pi$'s being coplanar. This weight is another connection between the optimization and the network, besides the predicted patch pairs themselves.
It is computed based on the feature distance of two patches, denoted by $d_\text{f}(p,q)$, extracted by the network:
$
w_{(p,q)} = e^{-d_\text{f}^2(p,q) / (\sigma^2 d_\text{fm}^2)},
$
where $d_\text{fm}$ is the maximum feature distance and $\sigma=0.6$.

The \emph{coplanarity regularization term} is defined as:
\begin{equation}\small
E_{\text{reg-cop}}(s) = \sum_{(i,j)\in \mathcal{P}}\sum_{\pi \in \Pi_{ij}}{\mu\, w_\pi\, \Psi(s_\pi)},
\label{eq:regcop}
\end{equation}
where the penalty function is defined as $\Psi(s) = \left(\sqrt{s}-1\right)^2$. Intuitively, minimizing this term together with the data term encourages the selection of pairs incurring a small value for the data term, while immediately pruning those pairs whose data term value is too large and deemed to be hard to minimize.  $w_\pi$ is defined the same as before, and
$\mu$ is a weighting variable that controls the emphasis on pair selection.

The \emph{key-point data term} is defined as:
\begin{equation}\small
E_{\text{data-kp}}({T},s)
= \sum_{(i,j) \in \mathcal{P}}\sum_{\theta \in \Theta_{ij}}{s_\theta\, ||\mathbf{T}_i \mathbf{u} - \mathbf{T}_j \mathbf{v}||},
\label{eq:datatermkp}
\end{equation}
Similar to coplanarity,
a control variable $s_\theta$ is used to determine the selection of point pair $\theta$, subjecting to the
\emph{key-point regularization term}:
\begin{equation}\small
E_{\text{reg-kp}}(s) = \sum_{(i,j)\in \mathcal{P}}\sum_{\theta \in \Theta_{ij}}{\mu\, \Psi(s_\theta)},
\label{eq:regkp}
\end{equation}
where $\mu$ shares the same weighting variable with Equation (\ref{eq:regcop}).

\vspace*{2mm}\noindent{\bf Optimization:}
The optimization of Equation (\ref{eq:robustopt}) is conducted iteratively, where
each iteration interleaves the optimization of transformations $T$ and selection variables $s$.
Ideally, the optimization could take every pair of frames in a sequence as an input for global optimization.
However, this is prohibitively expensive since for each frame pair the system
scales with the number of patch pairs and key-point pairs.
To alleviate this problem, we split the sequence into a list of overlapping fragments,
optimize frame poses within each fragment, and then perform a final global registration of the fragments, as in \cite{choi2015robust}.

For each fragment, the optimization takes all frame pairs within that fragment and registers them into a rigid point cloud. After that, we take the matching pairs that have been selected by the intra-fragment optimization, and solve the inter-fragment registration based on those pairs. Inter-fragment registration benefits more from long-range coplanarity predictions.

The putative matches found in this manner are then pruned further with a rapid and approximate RANSAC algorithm applied for each pair of fragments.   Given a pair of fragments, we randomly select a set of three matching feature pairs, which could be either planar-patch or key-point pairs.   We compute the transformation aligning the selected triplet, and then estimate the ``support'' for the transformation by
counting the number of putative match pairs that are aligned by the transformation. For patch pairs, alignment error is measures by the root-mean-square closest distance between sample points on the two patches. For key-point pairs, we simply use the Euclidean distance. Both use the same threshold of $1$cm.  If a transformation is found to be sufficiently supported by the matching pairs (more than $25\%$ consensus), we include all the supporting pairs into the global optimization.
Otherwise, we simply discard all putative matches. 

Once a set of pairwise constraints have been established in this manner, the frame transformations and pair selection variables are alternately optimized with an iterative process using Ceres \cite{agarwal2012ceres} for the minimization of the objective function at each iteration.
The iterative optimization converges when the relative value change of each unknown is
less than $1 \times 10^{-6}$.
At a convergence, the weighting variable $\mu$, which was initialized to $1$m in the beginning,
is decreased by half and the above iterative optimization continues.
The whole process is repeated until $\mu$ is lower than $0.01$m, which usually takes less than $50$ iterations.
The supplementary material provides a study of the optimization
behavior, including convergence and robustness to incorrect pairs.

\section{Results and Evaluations}
\label{sec:result}

\subsection{Training Set, Parameters, and Timings}
Our training data is generated from the ScanNet~\cite{dai2017scannet} dataset, which contains $1513$
scanned sequences of indoor scenes, reconstructed by BundleFusion~\cite{dai2017bundlefusion}.
We adopt the training/testing split provided with ScanNet and
the training set ($1045$ scenes) are used to generate our training triplets.
Each training scene contributes $10K$ triplets.
About $10M$ triplets in total are generated from all training scenes.
For evaluating our network, we build a \emph{coplanarity benchmark}
using $100$ scenes from the testing set.
For hierarchical optimization, the fragment size is $21$, with a $5$-frame overlap between adjacent fragments.
The network training takes about $20$ hours to converge.
For a sequence of $1K$ frames with $62$ fragments and $30$ patches per frame,
the running time is $10$ minutes for coplanarity prediction ($0.1$ second per patch pair)
and $20$ minutes for optimization ($5$ minutes for intra-fragment and $15$ minutes for inter-fragment).

\subsection{Coplanarity Benchmark}
We create a benchmark \textbf{COP} for evaluating RGB-D-based coplanarity matching of planar patches.
The benchmark dataset contains $12K$ patch pairs with ground-truth coplanarity, which
are organized according to the physical size/area of patches (COP-S)
and the centroid distance between pairs of patches (COP-D).
COP-S contains $6K$ patch pairs which are split uniformly into three subsets
with \emph{decreasing} average patch size,
where the patch pairs are sampled at random distances.
COP-D comprises three subsets (each containing $2K$ pairs)
with \emph{increasing} average pair distance but uniformly distributed patch size.
For all subsets, the numbers of positive and negative pairs are equal.
Details of the benchmark are provided in the supplementary material.

\subsection{Network Evaluation}
Our network is the first, to our knowledge, that is trained for coplanarity prediction.
Therefore, we perform comparison against baselines and ablation studies.
See visual results of coplanarity matching in the supplementary material.

\begin{figure}[t!] \centering
    \begin{overpic}[width=0.8\linewidth]{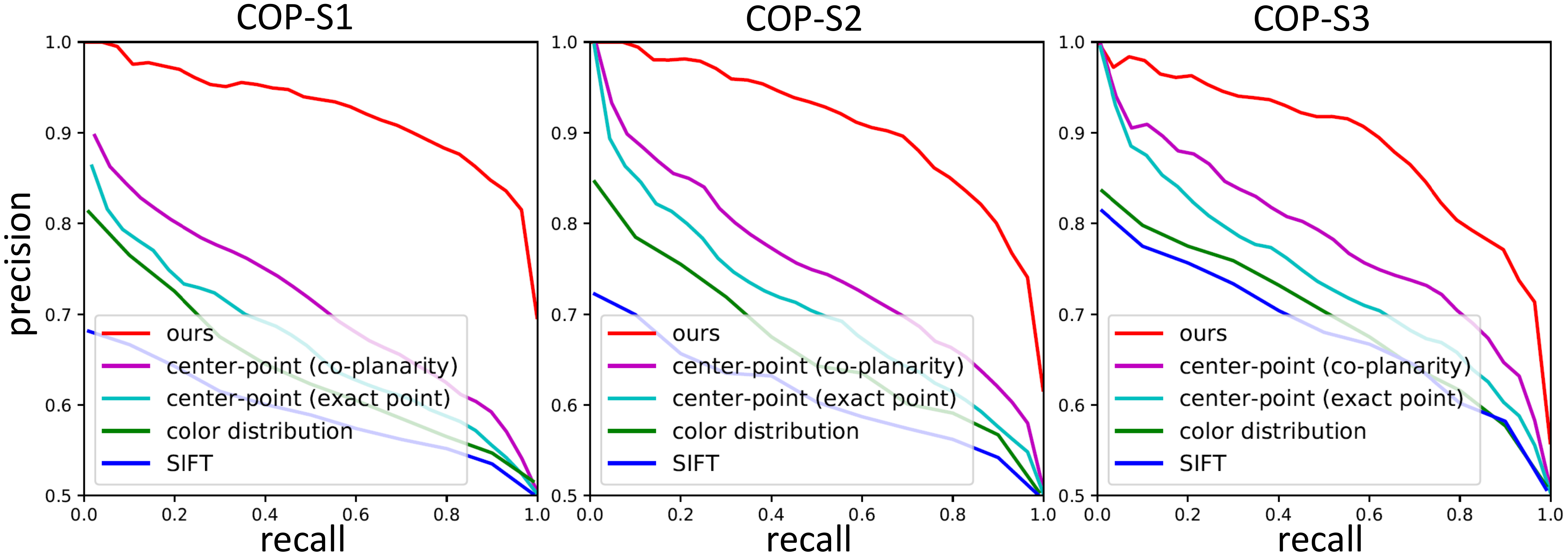}
    \end{overpic}
    \caption{Comparing to baselines including center-point matching
    networks trained with coplanarity and exact point matching, respectively,
    SIFT-based point matching and color distribution based patch matching.}
    \label{fig:plotbaseline}
\end{figure} 

\vspace*{2mm}\noindent{\bf Comparing to Baseline Methods:}
We first compare to two hand-crafted descriptors, namely
the color histogram within the patch region and the SIFT feature at the patch centroid.
%
For the task of key-point matching, a commonly practiced method (e.g., in~\cite{chang2017matterport})
is to train a neural network that takes image patches centered around the key-points as input.
We extend this network to the task of coplanarity prediction, as a non-trivial baseline.
For a fair comparison, we train a triplet network with ResNet-50 with only one tower per
patch taking three channels (RGB, depth, and normal) as input.
For each channel, the image is cropped around the patch centroid,
with the same padding and resizing scheme as before.
Thus, no mask is needed since the target is always at the image center.
We train two networks with different triplets for the task of
1) exact center point matching and
2) coplanarity patch matching, respectively.

The comparison is conducted over COP-S and the results of precision-recall are plotted in Figure~\ref{fig:plotbaseline}.
The hand-crafted descriptors fail on all tests, which shows the difficulty of our benchmark datasets.
Compared to the two alternative center-point-based networks
(point matching and coplanarity matching), our method performs significantly better,
especially on larger patches.

\begin{figure}[t!] \centering
    \begin{overpic}[width=0.8\linewidth]{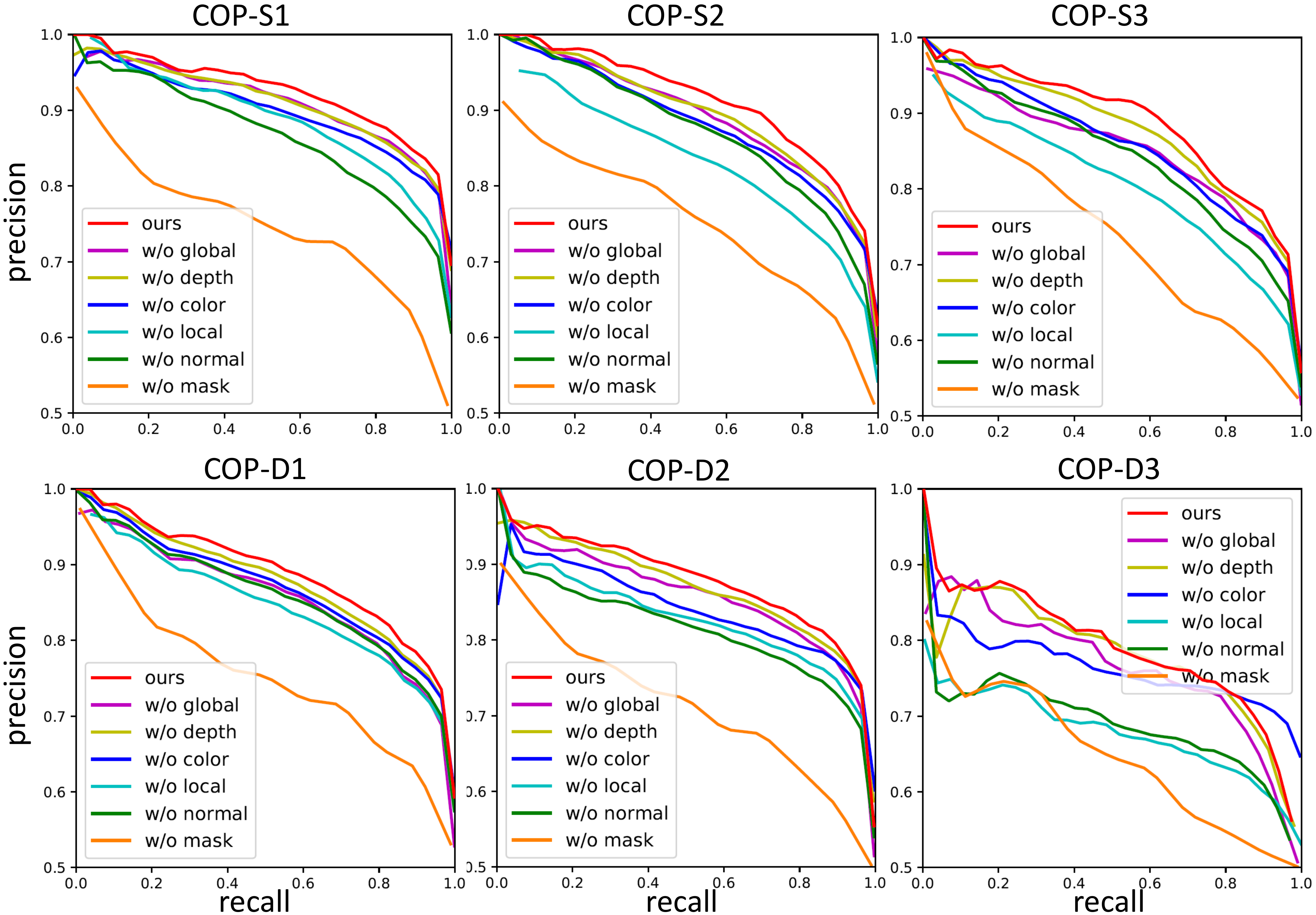}
    \end{overpic}
    \caption{Ablation studies of our coplanarity network.}
    \label{fig:plotablation}
\end{figure} 

\vspace*{2mm}\noindent{\bf Ablation Studies:}
To investigate the need for the various input channels,
we compare our full method against that with the RGB, depth, normal, or mask input disabled,
over the COP benchmark.
To evaluate the effect of multi-scale context, our method
is also compared to that without local or global channels.
The PR plots in Figure~\ref{fig:plotablation}
show that our full method works the best for all tests.

From the experiments, several interesting phenomena can be observed.
First, the order of overall importance of the different channels is:
mask $>$ normal $>$ RGB $>$ depth.
This clearly shows that coplanarity prediction across different views can neither rely on appearance
or geometry alone.
The important role of masking in concentrating the network's attention is quite evident.
We provide a further comparison to justify our specific masking scheme
in the supplementary material.
Second, the global scale is more effective for bigger patches and more distant pairs, for which
the larger scale is required to encode more context.
The opposite goes for the local scale due the higher resolution of its input channels.
This verifies the complementary effect of the local and global channels in capturing contextual information at different scales.

\subsection{Reconstruction Evaluation}

\noindent{\bf Quantitative Results:}
We perform a quantitative evaluation of reconstruction
using the TUM RGB-D dataset by~\cite{sturm12iros},
for which ground-truth camera trajectories are available.
Reconstruction error is measured by the absolute trajectory error (ATE), i.e.,
the root-mean-square error (RMSE) of camera positions along a trajectory.
%
%
We compare our method with six state-of-the-art reconstruction methods, including
RGB-D SLAM~\cite{endres2012evaluation},
VoxelHashing~\cite{niessner2013hashing},
ElasticFusion~\cite{Whelan15rss},
Redwood~\cite{choi2015robust},
BundleFusion~\cite{dai2017bundlefusion}, and
Fine-to-Coarse~\cite{halber2016fine}.
Note that unlike the other methods, Redwood does not use color information.
Fine-to-Coarse is the most closely related to our method, since it uses planar surfaces
for structurally-constrained registration. This method, however, relies on a good initialization
of camera trajectory to bootstrap, while our method does not.
Our method uses SIFT features for key-point detection and matching.
We also implement an enhanced version of our method where the key-point matchings are pre-filtered by BundleFusion (named `BundleFusion+Ours').

%
As an ablation study, we implement five baseline variants of our method.
1) `Coplanarity' is our optimization with only coplanarity constraints.
Without key-point matching constraint, our optimization can sometimes be under-determined and needs
reformulation to achieve robust registration when not all degrees of freedom (DoFs) can be fixed by coplanarity.
The details on the formulation can be found in the supplementary material.
2) `Keypoint' is our optimization with only SIFT key-point matching constraints.
3) `No D. in RANSAC' stands for our method where we did not use our learned patch descriptor
during the voting in frame-to-frame RANSAC.
In this case, any two patch pairs could cast a vote if they are geometrically aligned by the candidate transformation.
4) `No D. in Opt' means that the optimization objective for coplanarity is not weighted
by the matching confidence predicted by our network ($w_\pi$ in Equation (\ref{eq:dataterm}) and (\ref{eq:regcop})).
5) `No D. in Both' is a combination of 3) and 4).




\begin{table}[t]
\centering
\small
    \subfloat[Comparison to alternatives.]{%
    \scalebox{0.75}{
		\begin{tabular}{| c || c | c | c | c |}
			\hline
			Method & fr1/desk & fr2/xyz &  fr3/office & fr3/nst \\
            \hline
			RGB-D SLAM & 2.3& \blue{0.8} & 3.2 & 1.7  \\ \hline
			VoxelHashing & 2.3 & 2.2 & 2.3 & 8.7  \\ \hline
			Elastic Fusion & 2.0 & \green{1.1} &1.7& 1.6  \\ \hline
			Redwood & 2.7& 9.1 & 3.0& 192.9  \\ \hline
            Fine-to-Coarse  & 5.0& 3.0& 3.9& 3.0 \\ \hline
            BundleFusion  & 1.6& \green{1.1} & 2.2& \green{1.2} \\
            \hline\hline
            Ours & \green{1.4} & \green{1.1} & \green{1.6} & 1.5 \\
            \hline
            BundleFuison+Ours  & \blue{1.3} & \blue{0.8} & \blue{1.5} & \blue{0.9} \\
            \hline
		\end{tabular}
    }
    }
    \subfloat[Comparison to baselines.]{%
    \scalebox{0.75}{
       \begin{tabular}{| c || c | c | c | c |}
			\hline
			Method & fr1/desk & fr2/xyz &  fr3/office & fr3/nst \\
            \hline
			No D. in RANSAC & 9.6 & 4.8 & 12.6 & 2.3  \\ \hline
			No D. in Opt. & 4.8 & 2.7 & 2.5 & 1.9  \\ \hline
            No D. in Both  & 18.9 & 8.3 & 16.4 & 2.4 \\ \hline
			Key-point Only &5.6& 4.4 & 5.2 &2.6  \\ \hline
            Coplanarity Only & 2.5& 2.1& 3.7&--  \\
            \hline\hline
            Ours & \blue{1.4} & \blue{1.1} & \blue{1.7} & \blue{1.5} \\
            \hline
	   \end{tabular}
    }
    }
    \caption{Comparison of ATE RMSE (in cm) with alternative and baseline methods on TUM sequences~\protect\cite{sturm12iros}. Colors indicate the {\color{blue}best} and {\color{green}second best} results.}
    \label{table:comp_tum}
\end{table}

Table~\ref{table:comp_tum} reports the ATE RMSE comparison.
%
Our method achieves state-of-the-art results
for the first three TUM sequences (the fourth is a flat wall). This is achieved by exploiting our long-range coplanarity matching
for robust large-scale loop closure, while utilizing key-point based matching to pin down the possible free DoFs which are not determinable by coplanarity.
When being combined with BundleFusion key-points, our method achieves the best results over all sequences.
Therefore, our method complements the current state-of-the-art methods by providing a means
to handle limited frame-to-frame overlap.

The ablation study demonstrates the importance of our learned patch descriptor
in our optimization -- i.e., our method performs better than all variants that do not include it.
It also shows that coplanarity constraints alone are superior to keypoints only for all sequences except the flat wall (fr3/nst).   Using coplanar and keypoint matches together provides the best method overall.

\begin{figure*}[t!] \centering
    \begin{overpic}[width=0.999\linewidth]{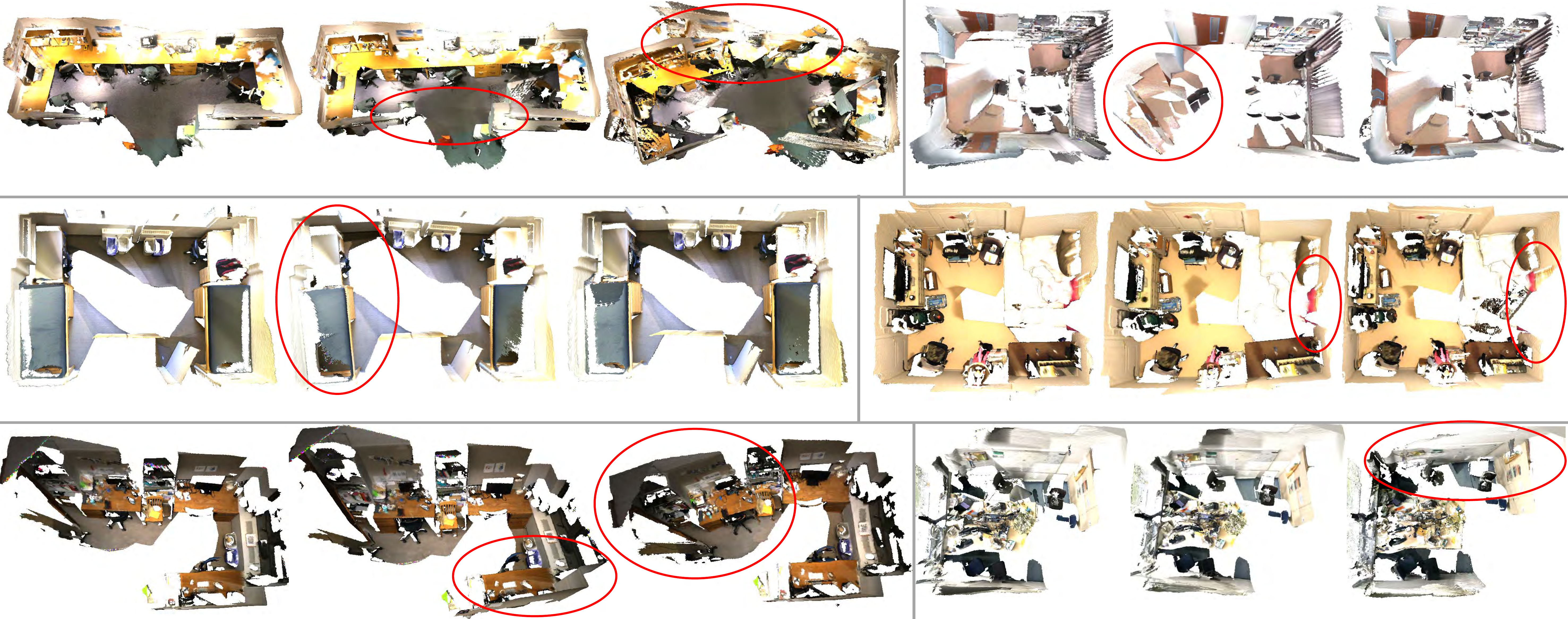}
    \scriptsize
        \put(5,27.3){Ours ({\color{red}$40660$})}
        \put(28,27.3){BF}
        \put(45,27.3){F2C}
        \put(60,27.3){Ours ({\color{red}$9607$})}
        \put(79,27.3){BF}
        \put(90,27.3){F2C}
        \put(5,13){Ours ({\color{red}$13791$})}
        \put(28,13){BF}
        \put(45,13){F2C}
        \put(58,13){Ours ({\color{red}$4975$})}
        \put(77,13){BF}
        \put(90,13){F2C}
        \put(5,-1.4){Ours ({\color{red}$2712$})}
        \put(28,-1.4){BF}
        \put(45,-1.4){F2C}
        \put(60,-0.7){Ours ({\color{red}$9889$})}
        \put(79,-0.7){BF}
        \put(90,-0.7){F2C}
    \end{overpic}
    \caption{Visual comparison of reconstructions by our method, BundleFusion (BF)~\protect~\cite{dai2017bundlefusion}, and Fine-to-Coarse (F2C)~\protect~\cite{halber2016fine}, on six sequences.
    Red ellipses indicate parts with misalignment.
    For our results, we give the number of long-range coplanar pairs selected by the optimization.
    }
    \vspace{-10pt}
    \label{fig:gallery}
\end{figure*}
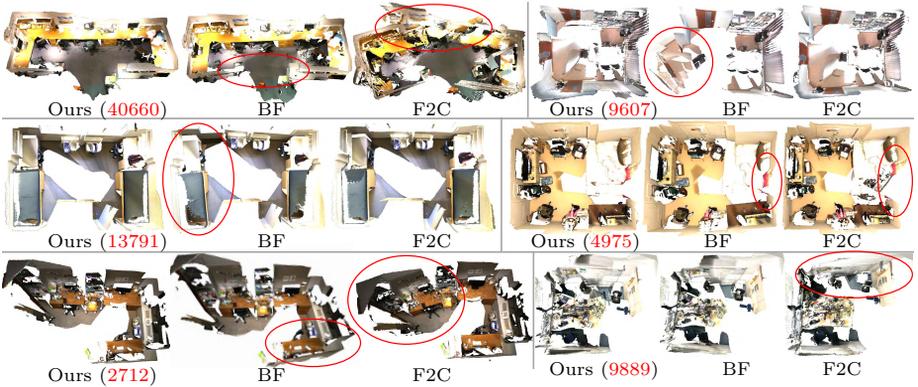

\vspace*{2mm}\noindent{\bf Qualitative Results:}
Figure~\ref{fig:gallery}
shows visual comparisons of reconstruction on sequences from ScanNet~\cite{dai2017scannet}
and new ones scanned by ourselves.
We compare reconstruction results of our method with
a state-of-the-art key-point based method (BundleFusion)
and a planar-structure-based method (Fine-to-Coarse).
The low frame overlap makes the key-point based loop-closure detection fail in BundleFusion.
Lost tracking of successive frames provides a poor initial alignment for Fine-to-Coarse, causing it to fail.
In contrast, our method can successfully detect non-overlapping loop closures through coplanar patch pairs
and achieve good quality reconstructions for these examples without an initial registration.
%
More visual results are shown in the supplementary material.

\begin{figure}[t!] \centering
    \begin{overpic}[width=0.88\linewidth]{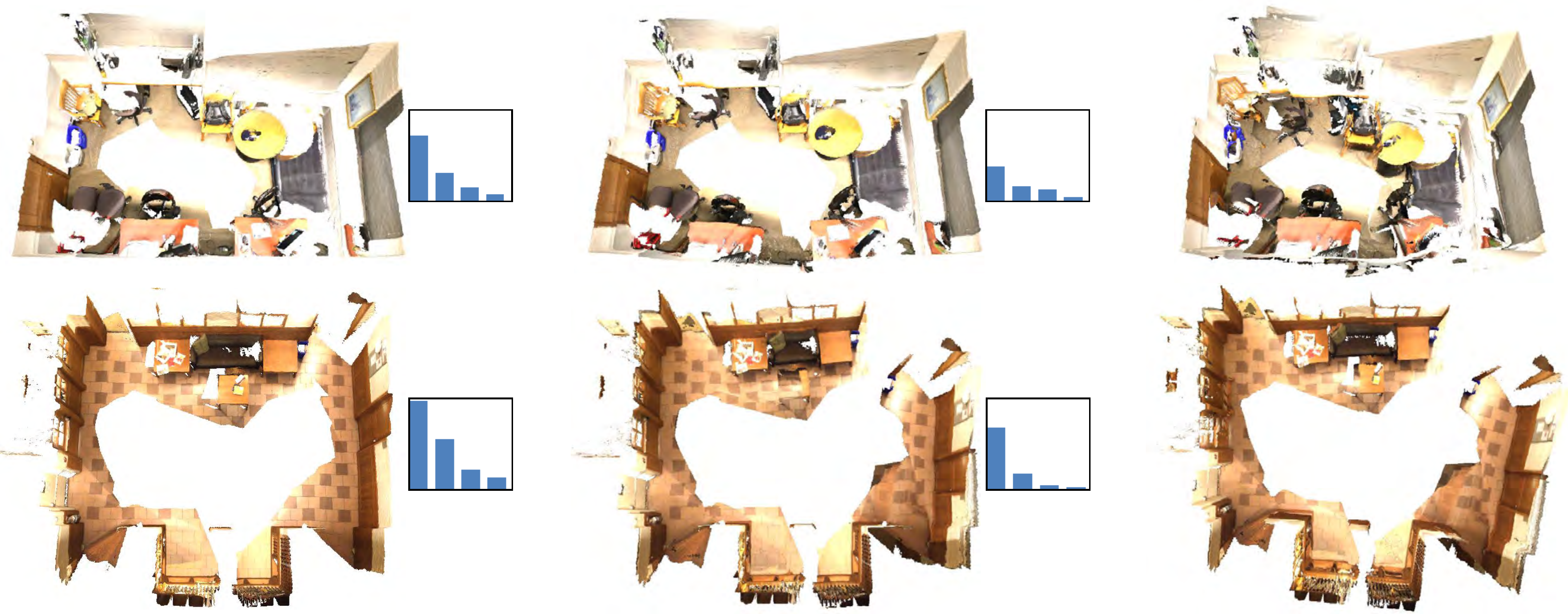}
    \end{overpic}
    \caption{Reconstruction results with $100\%$ (left column), $50\%$ (middle) and $0\%$ (right) of long-range coplanar pairs detected, respectively.
    The histograms of long-range coplanar patch pairs (count over patch distance ($1\sim5$m)) are given.
    }
    \label{fig:longrange}
    \vspace{-15pt}
\end{figure}

\vspace*{2mm}\noindent{\bf Effect of Long-Range Coplanarity.}
To evaluate the effect of long-range coplanarity matching on reconstruction quality,
we show in Figure~\ref{fig:longrange} the reconstruction results computed with all,
half, and none of the long-range coplanar pairs predicted by our network.
We also show a histogram of coplanar pairs survived the optimization.
From the visual reconstruction results, the benefit of long-range coplanar pairs is apparent.
In particular, the larger scene (bottom) benefits more from long-range coplanarity than the smaller one (top).
In Figure~\ref{fig:gallery}, we also give the number of non-overlapping coplanar pairs after optimization,
showing that long-range coplanarity did help in all examples.

\section{Conclusion}\vspace{-5pt}
We have proposed a new planar patch descriptor designed for finding coplanar patches without a priori global alignment.  At its heart, the method uses a deep network to map planar patch inputs with RGB, depth, and normals to a descriptor space where proximity can be used to predict coplanarity.
We expect that deep patch coplanarity prediction provides a useful complement to existing features for SLAM applications, especially in scans with large planar surfaces and little inter-frame overlap.

\vspace{-10pt}


\subsubsection{Acknowledgement}
We are grateful to Min Liu, Zhan Shi, Lintao Zheng, and Maciej Halber for their help on data preprocessing. We also thank Yizhong Zhang for the early discussions. This work was supported in part by the NSF (VEC 1539014/ 1539099, IIS 1421435, CHS 1617236), NSFC (61532003, 61572507, 61622212), Google, Intel, Pixar, Amazon, and Facebook. Yifei Shi was supported by the China Scholarship Council.

{\small
\bibliographystyle{splncs}
\bibliography{decop_eccv18}
}

\title{PlaneMatch: Patch Coplanarity Prediction for Robust RGB-D Reconstruction \\ Supplemental Material} 
\titlerunning{PlaneMatch}

\author{Yifei Shi\inst{1,2} \and
Kai Xu \inst{1,2} \and
Matthias Nie{\ss}ner\inst{3} \and
Szymon Rusinkiewicz\inst{1} \and
Thomas Funkhouser\inst{1,4}}
\authorrunning{Yifei Shi et al.}

\institute{Princeton University\and
National University of Defense Technology\and
Technical University of Munich\and
Google}

\maketitle



\section{Outline}
In this supplemental material, we provide the following additional information and results:
\begin{itemize}
\item Section~\ref{sec:benchmark} provides an overview of the dataset of our coplanarity benchmark (COP).
\item Section~\ref{sec:network} gives more evaluation results for our coplanarity network, including a comparison of different masking schemes (Section~\ref{subsec:mask}), evaluation on patch pairs proposed from real cases of scene reconstruction (Section~\ref{subsec:copreal}), and visual qualitative results of coplanarity matching (Section~\ref{subsec:copvisual}).
\item Section~\ref{sec:recon} provides more evaluations of the reconstruction algorithm. Specifically, we first evaluate the robustness of the registration against the initial ratio of incorrect pairs (Section~\ref{subsec:robust}). 
We then compare the reconstruction performance of two alternative optimization strategies
(Section~\ref{subsec:opt}).
Finally, we show more visual results of reconstructions for scenes from various datasets
(Section~\ref{subsec:reconvisual}).
\item Section~\ref{sec:limitation} discusses the limitations of our method.
\item Finally, Section~\ref{sec:coponly} provides the formulation for a variatnt of our method that only utilizes coplanarity constraints (Section~\ref{subsec:formulation}), the optimization procedure used for that variant (Section~\ref{subsec:optimization}), and the stability analysis used for achieving a robust optimization in that variant (Section~\ref{subsec:stability}).

\end{itemize}

\section{COP Benchmark Dataset}
\label{sec:benchmark}
\begin{figure*}[t!] \centering
    \begin{overpic}[width=0.96\linewidth]{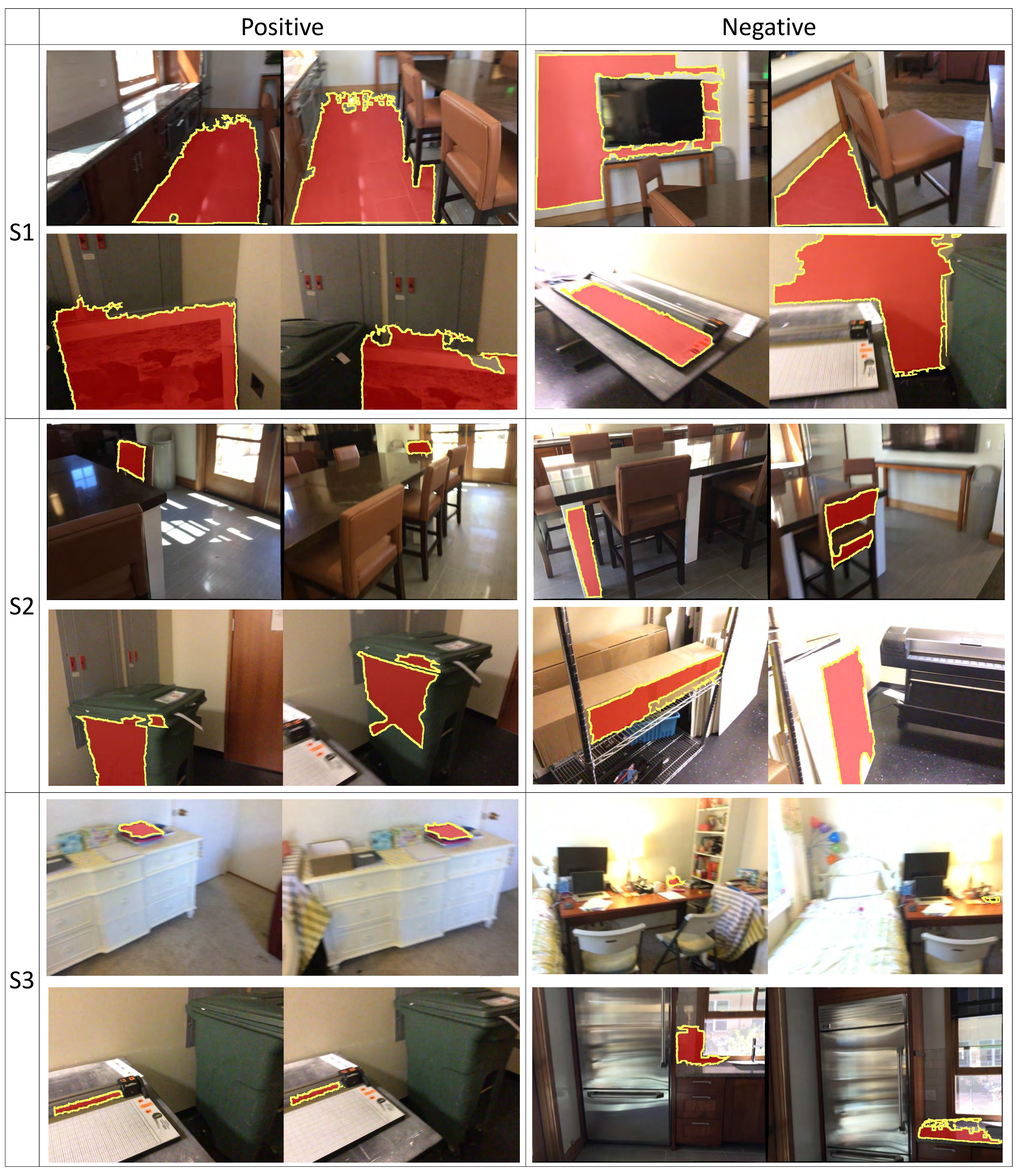}
    \small
    \end{overpic}
    \caption{An overview of the patch pairs (both positive and negative) in the benchmark dataset COP-S.
    The dataset is organized according to patch size.
    S1: $0.25$\textapprx$10$ m$^2$.
    S2: $0.05$\textapprx$0.25$ m$^2$.
    S3: $0$\textapprx$0.05$ m$^2$.
    }
    \label{fig:bmos}
\end{figure*}

\begin{figure*}[t!] \centering
    \begin{overpic}[width=0.96\linewidth]{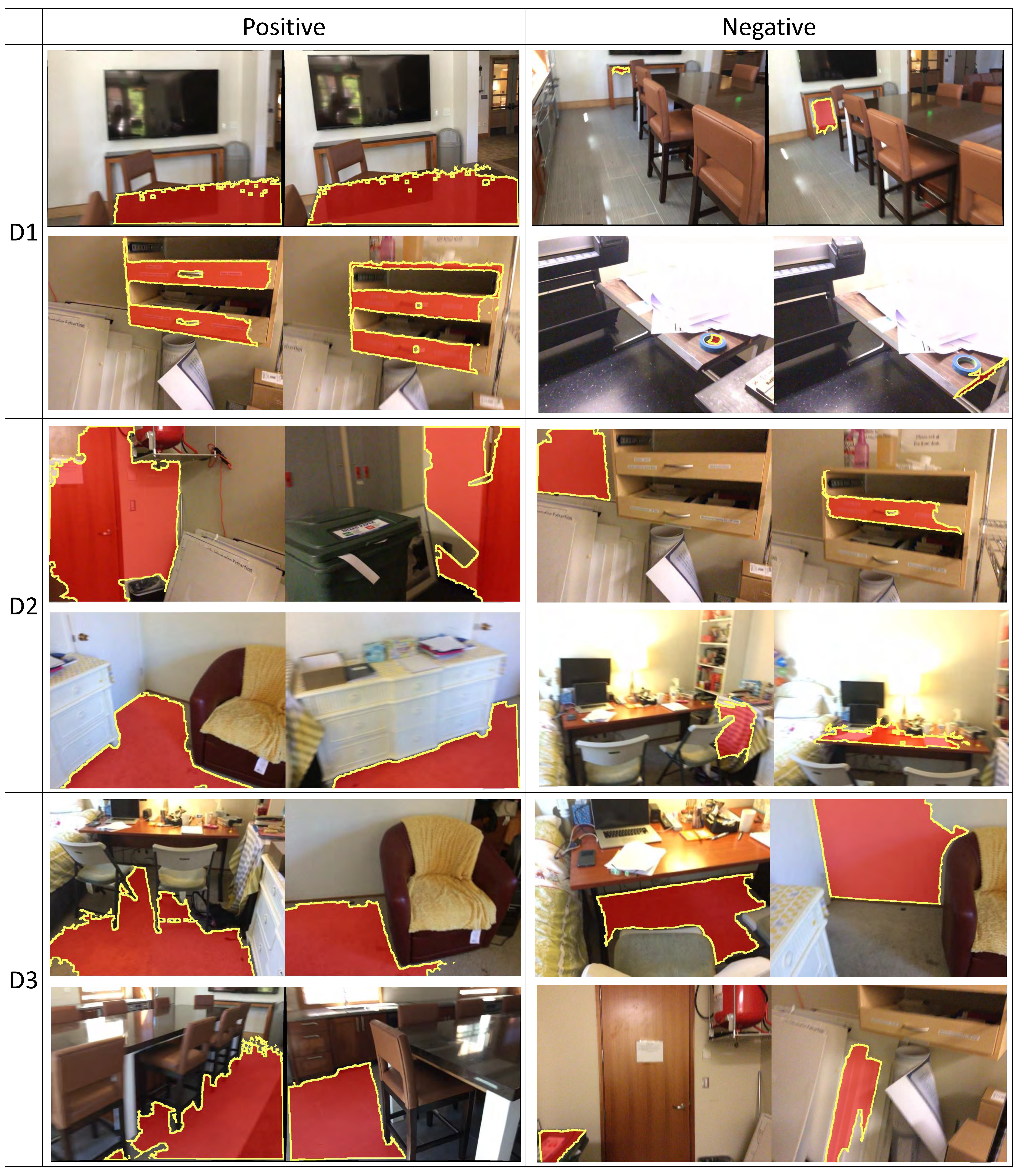}
    \small
    \end{overpic}
    \caption{An overview of the patch pairs (both positive and negative) in the benchmark dataset COP-D.
    The dataset is organized according to pair distance.
    D1: $0$\textapprx$0.3$ m.
    D2: $0.3$\textapprx$1$ m.
    D3: $1$\textapprx$5$ m.}
    \label{fig:bmod}
\end{figure*}

Figure~\ref{fig:bmos} and~\ref{fig:bmod} provide an overview of our coplanarity benchmark datasets, COP-S (organized
in decreasing patch size) and COP-D (in increasing patch distance), respectively.
For each subset, we show both positive and negative pairs, each with two pairs.
Note how non-trivial the negative pairs are in our dataset, for example, the negative pairs of S3 and D1.

\clearpage
\section{Network Evaluations}
\label{sec:network}

This section provides further studies and evaluations of the performance of our coplanarity prediction network.

\subsection{Different Masking Schemes}
\label{subsec:mask}
\begin{figure}[h!] \centering
    \begin{overpic}[width=0.7\linewidth]{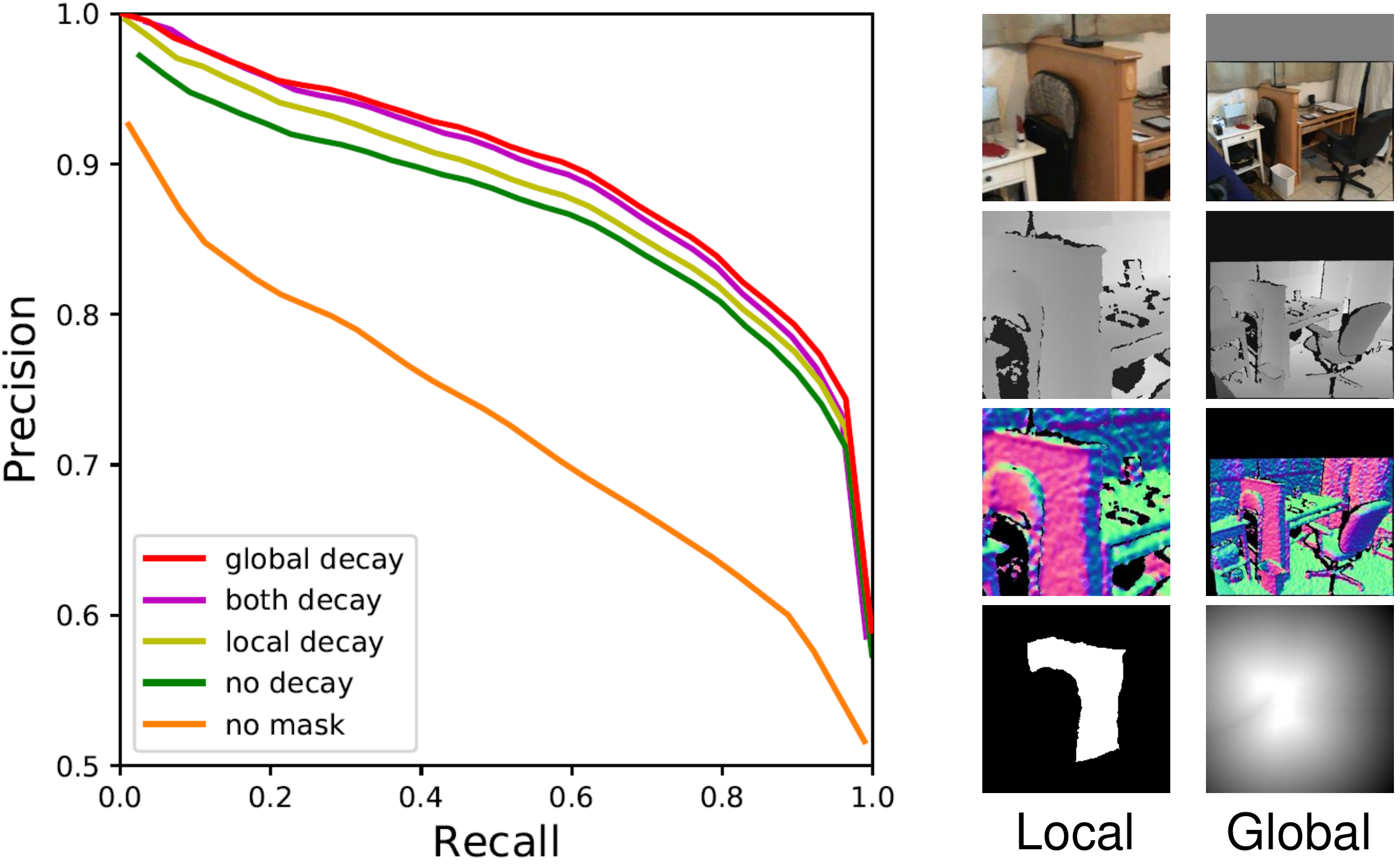}
    \end{overpic}
    \caption{Comparison of different masking schemes on the entire COP-D dataset. `Global decay' is our scheme.}
    \label{fig:mask_scheme}
\end{figure}

We first investigate several alternative masking schemes for the local and global inputs of our coplanarity network.
The proposed masking scheme is summarized as follows (see Figure~\ref{fig:mask_scheme} (right)).
The local mask is binary, with the patch of interest in white and the rest of the image in black.
The global mask, in contrast, is continuous, with the patch of interest in white and then
a smooth decay to black outside the patch boundary.

We compare in Figure~\ref{fig:mask_scheme} our masking scheme (global decay) with several alternatives including
1) using distance-based decaying for both local and global scale (both decay),
2) using distance-based decaying only for local scale (local decay),
3) without decaying for either scale (no decay),
and 4) without using a mask at all (no mask).
Over the entire COP-D benchmark dataset, we test the above methods and plot the PR curves.
The results demonstrate the advantage of our specific design choice of masking scheme
(using decaying for global scale but not for local).

\subsection{Performance on Patches Proposed during Reconstructions}
\label{subsec:copreal}

Our second study investigates the network performance for a realistic balance of positive and negative patch pairs.
The performance of our coplanarity network has so far been evaluated over the COP benchmark dataset, which contains comparable numbers of positive and negative examples.
To evaluate its performance in a reconstruction setting, we test on patch pairs proposed during the reconstruction
of two scenes (the full sequence of `fr1/desk' and `fr2/xyz' from the TUM dataset).
The ground-truth coplanarity matching is detected based on the ground-truth alignment provided with
the TUM dataset.

\begin{figure}[h!] \centering
    \begin{overpic}[width=0.96\linewidth]{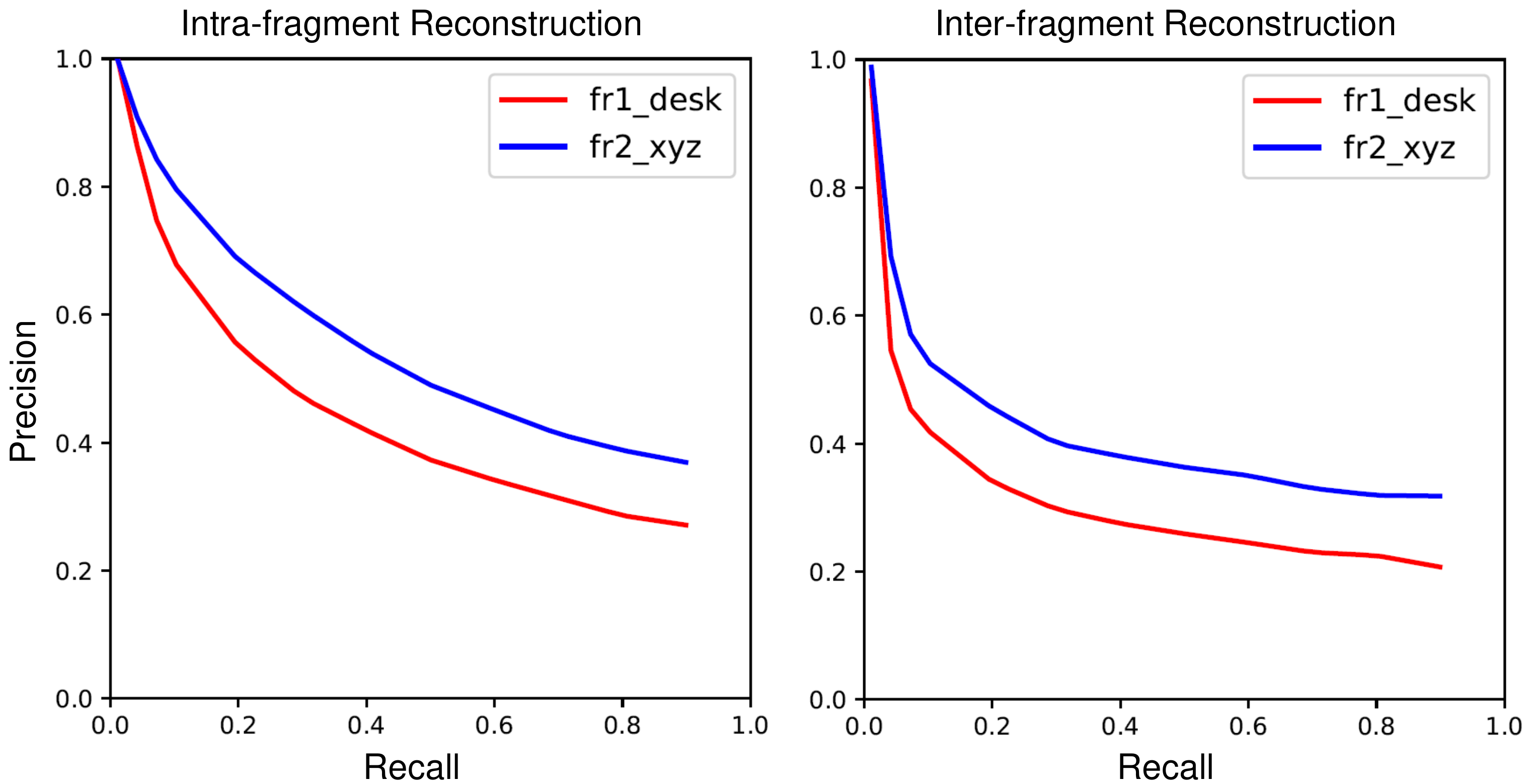}
    \end{overpic}
    \caption{Performance of our coplanarity network on patch pairs proposed from the reconstruction of
    sequences `fr1/desk' and `fr2/xyz' from the TUM dataset. The PR curves for both intra- (left) and inter-fragment
    (right) reconstruction are shown.}
    \label{fig:real_perf}
\end{figure}

Figure~\ref{fig:real_perf} shows the plot of PR curves for both intra- and inter-fragment
reconstructions. The values for intra-fragment are averaged over all fragments.
For patches from the real case of
scene reconstruction, our network achieves a precision of $>20\%$, when the recall rate is $80\%$.
This accuracy is sufficient for our robust optimization for frame registration, which can be seen
from the evaluation in Figure~\ref{fig:opt_robust}; see Section~\ref{subsec:robust}.

\subsection{More Visual Results of Coplanarity Matching}
\label{subsec:copvisual}
\begin{figure*}[t!] \centering
    \begin{overpic}[width=0.999\linewidth]{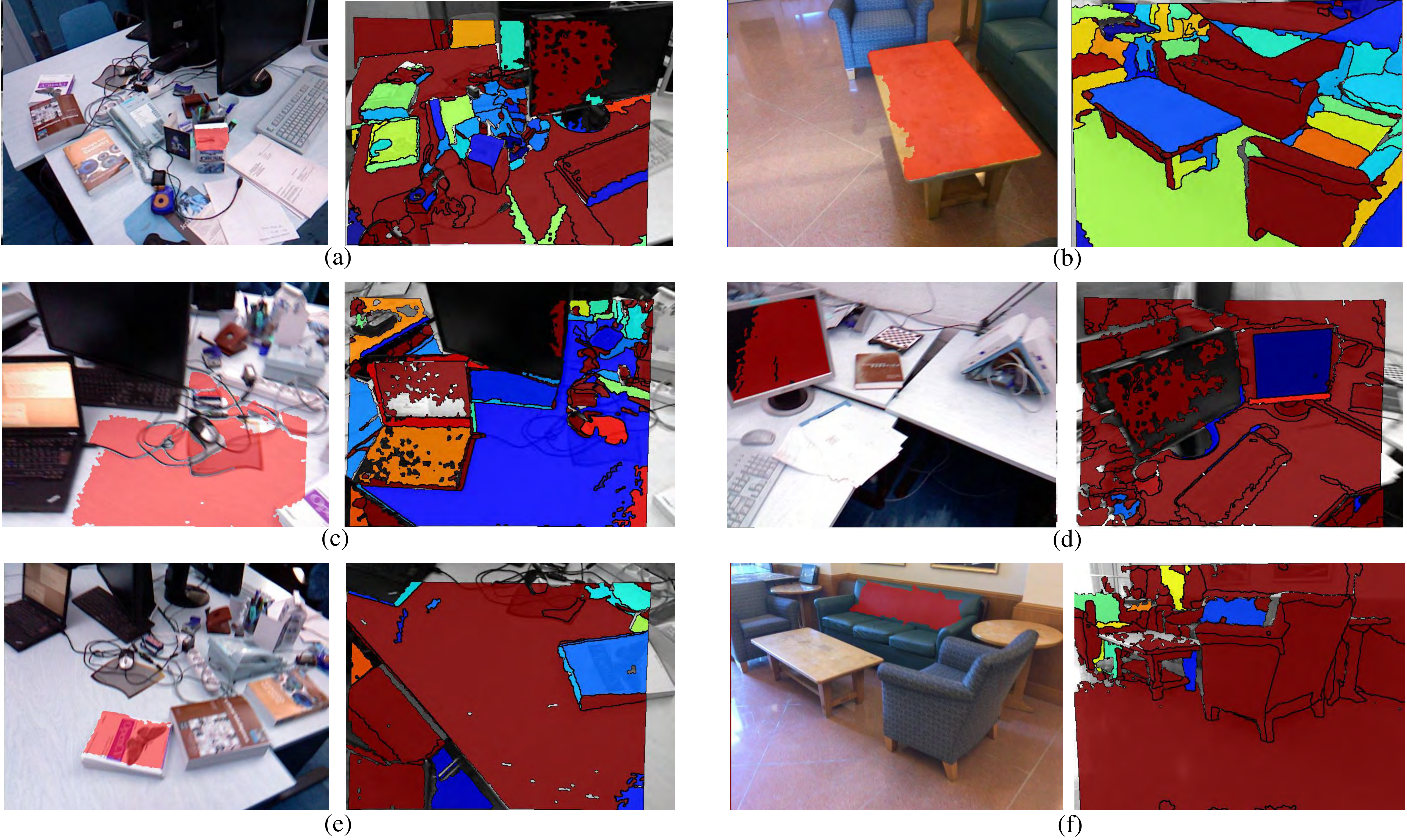}
    \small
    \end{overpic}
    \caption{Visualization of coplanarity matching for six query patches.
    For each example, the query patch is selected in the left image. In the right image,
    all patches are color-coded with the dissimilarity predicted by our coplanarity network (blue is small and red is large).}
    \label{fig:copvis}
\end{figure*}

Figure~\ref{fig:copvis} shows some visual results of coplanarity matching.
Given a query patch in one frame, we show all patches in another frame, which are color-coded
with the dissimilarity predicted by our coplanarity network (blue is small and red is large).
The results show that our network produces correct coplanarity embedding, even for patches
observed across many views.

\section{Reconstruction Evaluations}
\label{sec:recon}

\subsection{Robustness to Initial Coplanarity Accuracy}
\label{subsec:robust}
\begin{figure}[h] \centering
    \begin{overpic}[width=0.95\linewidth]{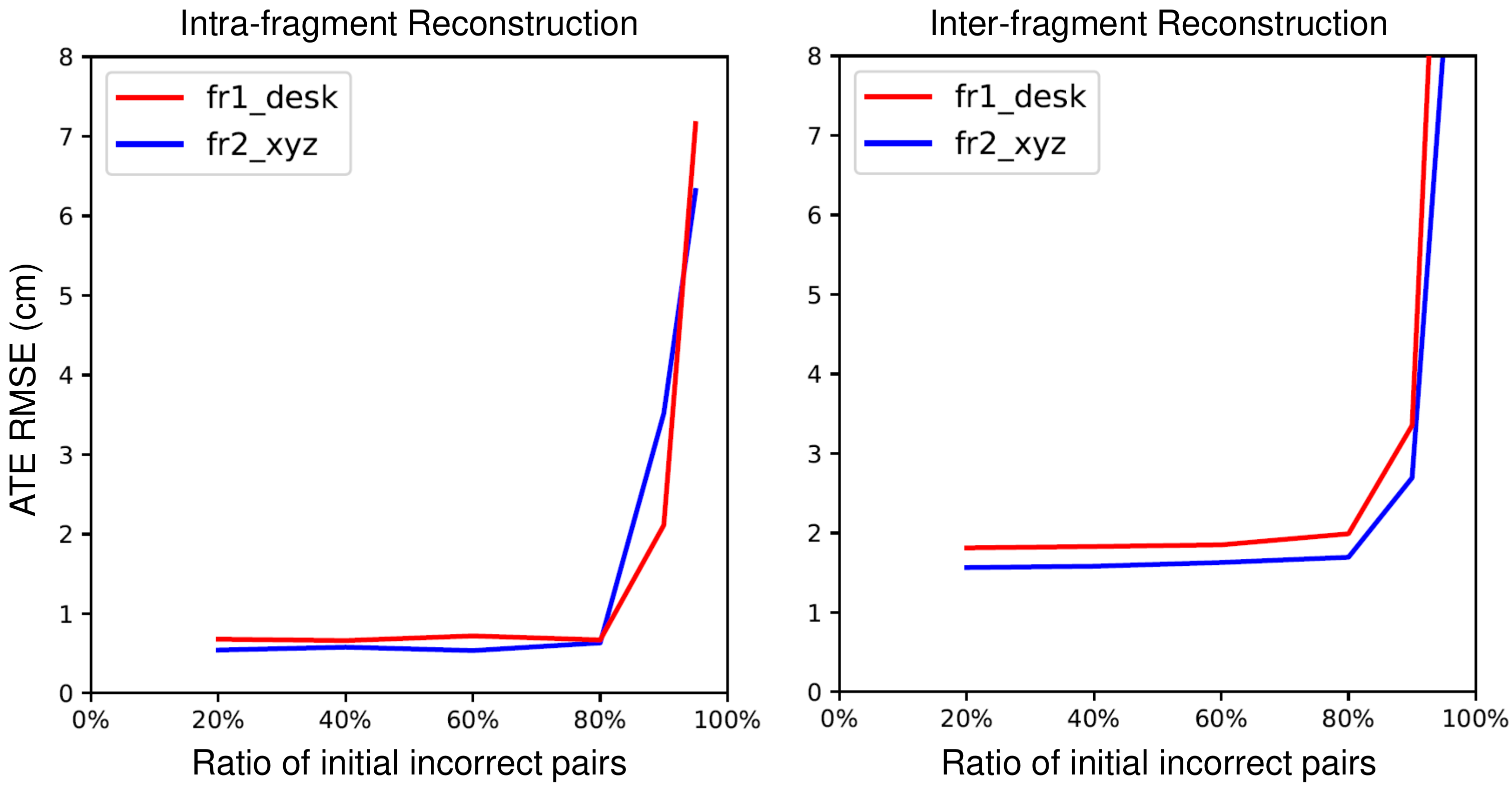}
    \end{overpic}
    \caption{Evaluation of the robustness of our coplanarity-based alignment on
    sequences `fr1/desk' and `fr2/xyz' from the TUM dataset. The plots shows the ATE RMSE (in cm)
    over different precisions. The results for both intra- (left) and inter-fragment
    (right) reconstruction are shown.}
    \label{fig:opt_robust}
\end{figure}

To evaluate the robustness of our optimization for coplanarity-based alignment,
we inspect how tolerant the optimization is to the initial accuracy of the coplanarity prediction.
In Figure~\ref{fig:opt_robust}, we plot the reconstruction error of our method on two
sequences (full) from TUM dataset, with varying ratio of incorrect input pairs.
In our method, given a pair of patches, if their feature distance in the embedding space is smaller than $2.5$,
it is used as a hypothetical coplanar pair being input to the optimization.
The varying incorrect ratios are thus obtained via gradually introducing more incorrect predictions
by adjusting the feature distance threshold.

Reconstruction error is measured by the absolute trajectory error (ATE), i.e., the root-mean-square error (RMSE) of camera positions along a trajectory.
The results demonstrate that our method is quite robust against the initial precision of coplanarity matching,
for both intra- and inter-fragment reconstructions.
In particular, the experiments show that our method is robust for a precision $20\%$ (incorrect ratio of $80\%$),
while keeping the recall rate no lower than $80\%$.

\subsection{Performance of Different Optimization Strategies}
\label{subsec:opt}
\begin{figure}[h!] \centering
    \begin{overpic}[width=0.5\linewidth]{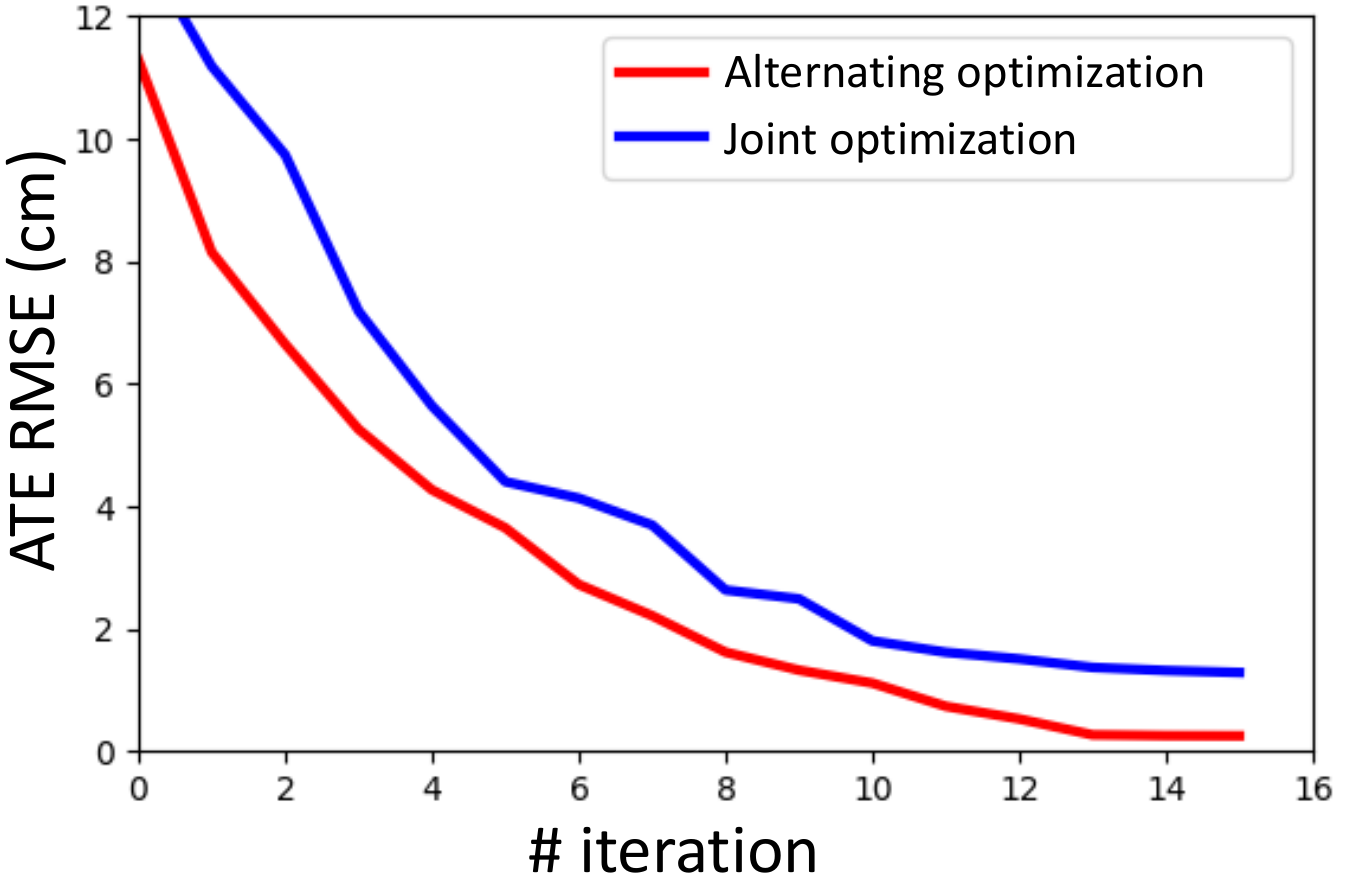}
    \end{overpic}
    \caption{Evaluation of different optimization strategies.}
    \label{fig:optimization_strategy}
\end{figure}

An alternative strategy for solving Equation (2) is to optimize transformations $T$ and selection
variables $s$ jointly.
In Figure~\ref{fig:optimization_strategy}, we report the reconstruction error over iterations on two sequences ('fr1/desk' and 'fr2/xyz') in the TUM dataset, for both alternating optimization and joint optimization.
The reconstruction error is calculated by averaging the ATE of the intra-fragment reconstructions and the inter-fragment reconstructions. The result shows that alternating optimization achieves better convergence performance. Similar results can be observed on other sequences from the TUM dataset.

\subsection{More Visual Results of Reconstruction}
\label{subsec:reconvisual}
\begin{figure*}[h!] \centering
    \begin{overpic}[width=0.96\linewidth]{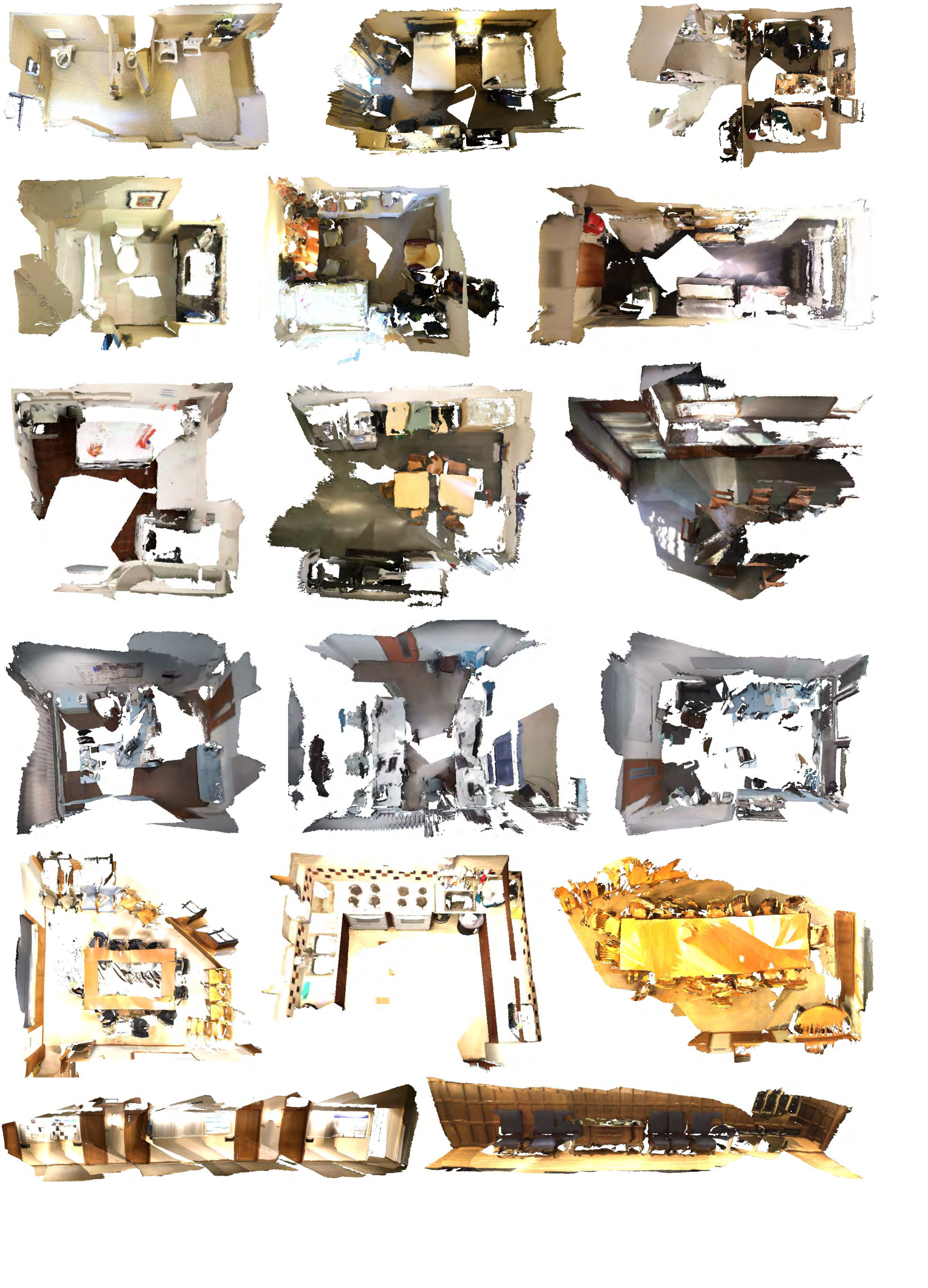}
    \small
    \end{overpic}
    \caption{Reconstruction results on $17$ sequences, including $9$ from ScanNet~\cite{dai2017scannet} (first three rows) and $8$ scanned by ourselves (last three rows).
    }
    \label{fig:recon}
\end{figure*}

\begin{figure*}[h!] \centering
    \begin{overpic}[width=0.8\linewidth]{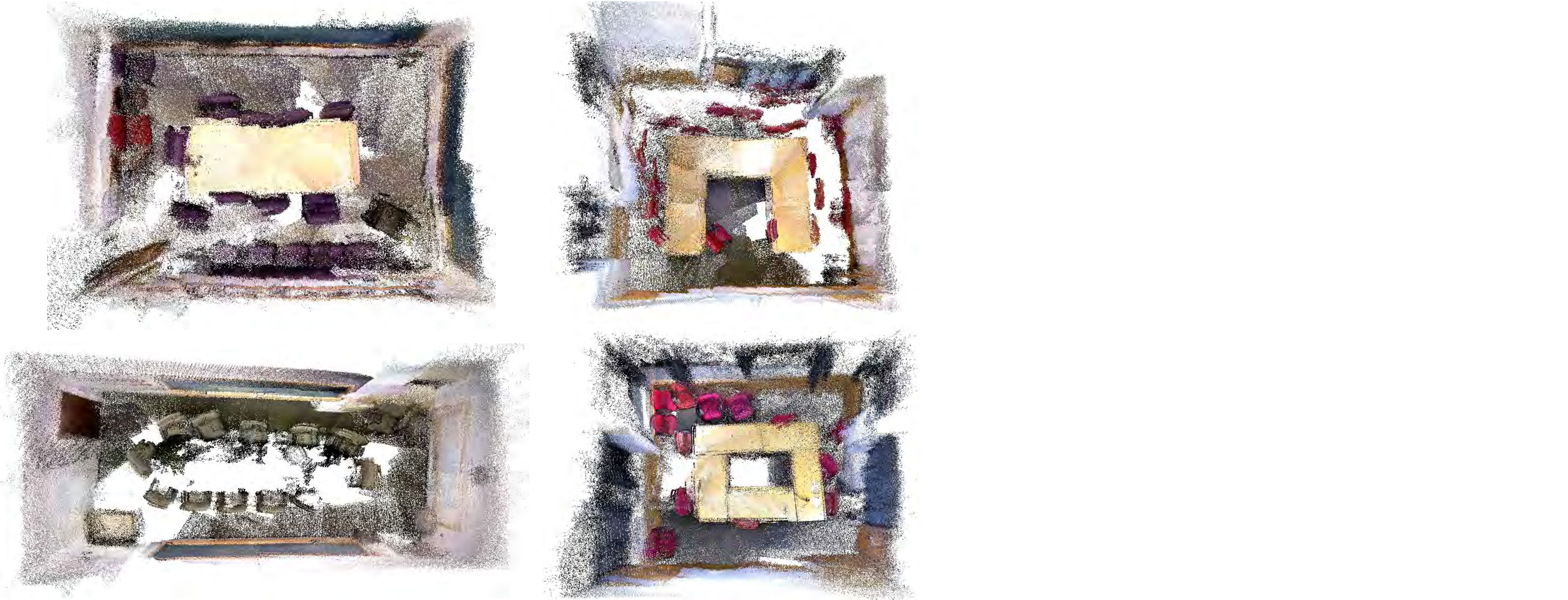}
    \small
    \end{overpic}
    \caption{Reconstruction results of four sequences from the Sun3D dataset~\cite{xiao2013sun3d}.
    }
    \label{fig:reconsun3d}
\end{figure*}

\begin{figure*}[h!] \centering
    \begin{overpic}[width=0.9\linewidth]{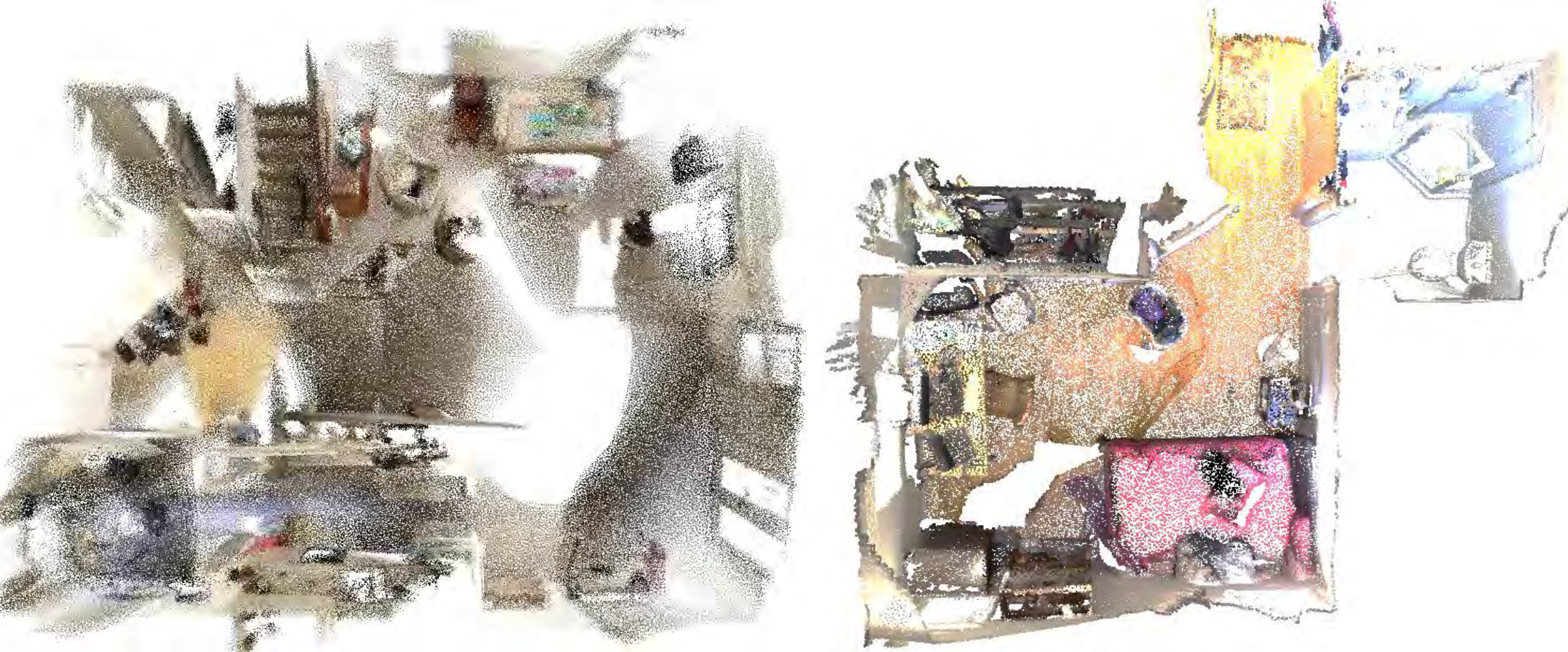}
    \small
    \end{overpic}
    \caption{Reconstruction results of two scenes with multiple rooms.}
    \label{fig:reconmultipleroom}
\end{figure*}

Figure~\ref{fig:recon} shows more visual results of reconstruction on $17$ sequences, including $9$ from the ScanNet dataset~\cite{dai2017scannet} and $8$ new ones scanned by ourselves.
The sequences scanned by ourselves have very sparse loop closure due the missing parts.
Our method works well for all these examples.
Figure~\ref{fig:reconsun3d} shows the reconstruction of $4$ sequences from the Sun3D dataset~\cite{xiao2013sun3d}.
Since the registration of Sun3D sequences is typically shown without fusion in previous works (e.g.,~\cite{xiao2013sun3d,halber2016fine}), we only show the point clouds.
Figure~\ref{fig:reconmultipleroom} shows two reconstruction examples on scene contains multiple rooms scanned by ourselves.


\section{Limitations, Failure Cases and Future work}
\label{sec:limitation}
Our work has several limitations, which suggest topics for future research.

\emph{First}, coplanarity correspondences alone are not always enough to constrain camera poses uniquely in some environments -- e.g., the pose of a camera viewing only a single flat wall will be under-constrained.
Therefore, coplanarity is \emph{not} a replacement for traditional features,
such as key-points, lines, etc.;
rather, we argue that coplanarity constraints provide additional signal and constraints which are critical in many scanning scenarios, thus helping to improve the reconstruction results.
This becomes particularly obvious in scans with a sparse temporal sampling of frames.

\begin{figure}[h!] \centering
    \begin{overpic}[width=0.7\linewidth]{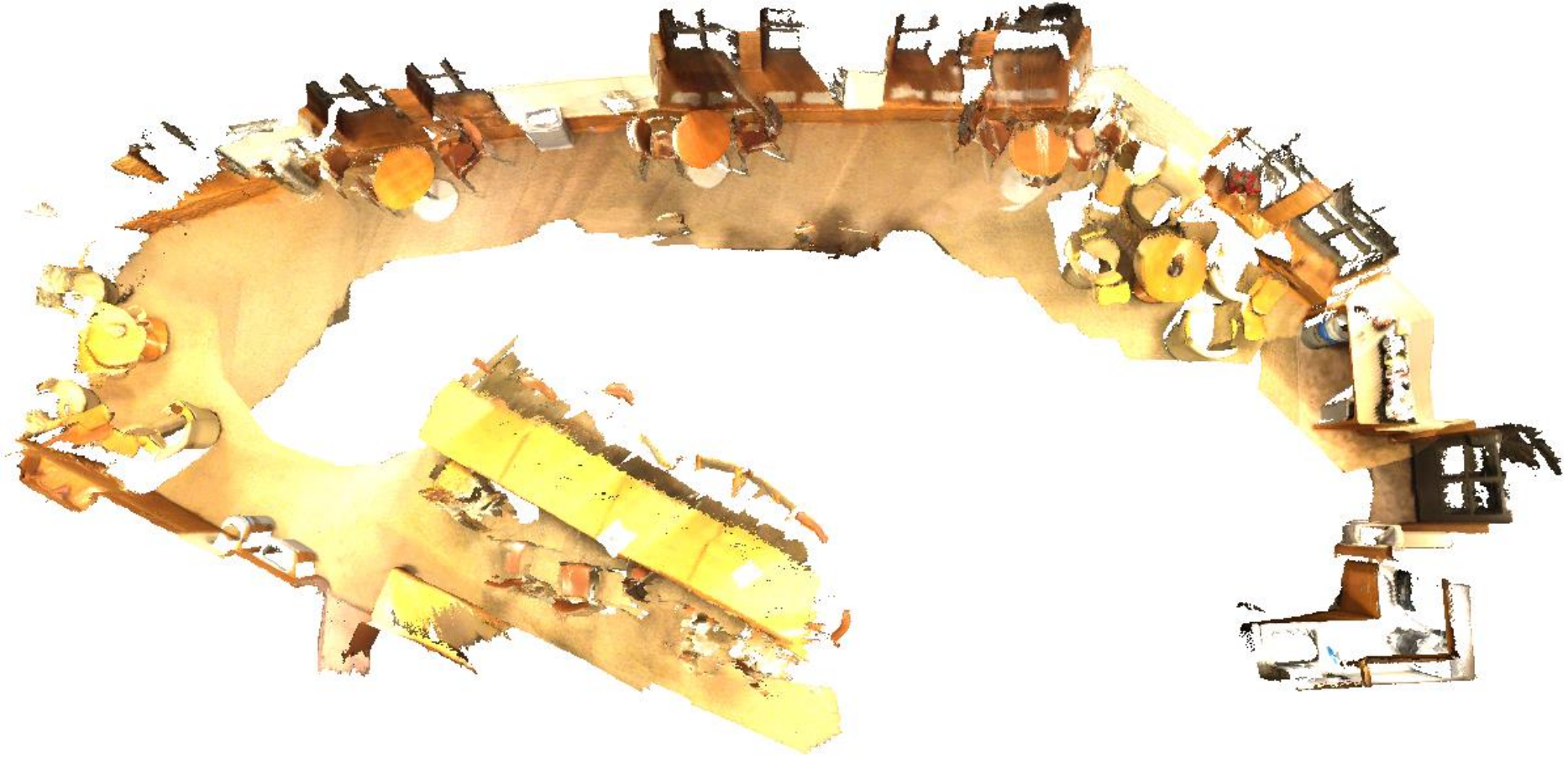}
    \end{overpic}
    \caption{
    The coplanarity constraint could cause over-regularization: A curvy wall (top) is mistakenly flattened causing the loop cannot be closed at the bottom wall for which long-range coplanarity is not available.}
    \label{fig:failure}
\end{figure}

\emph{Second},
for the cases where short-range coplanar patches dominate long-range ones
(e.g., a bending wall), our method
could reconstruct an overly flat surface due to the coplanarity regularization by
false positive coplanar patch pairs between adjacent frames.
For example, in Figure~\ref{fig:failure}, we show a tea room scanned by ourselves.
The top wall is not flat, but the false positive coplanar pairs detected between adjacent frames
could over-regularize the registration,
making it mistakenly flattened. This in turn causes the loop cannot be closed
at the wall in the bottom.

\emph{Third},
since the network prediction relies on the information of color, depth and normal, the prediction results could be wrong when the inputs fail to provide sufficient information on coplanarity. For example, two white walls without any context will always be predicted to be coplanar although they could be not.

\emph{Last}, our optimization is currently a computational bottleneck -- it takes approximately $20$ minutes to perform the robust optimization in typical scans shown in the paper.
Besides exploiting the highly parallelizable intra-fragment registrations,
a more efficient optimization is a worthy direction for future investigation. 
We are also interested in testing our method on a broader range of RGB-D datasets (e.g. the dataset in~\cite{park2017colored}).

\clearpage
\section{Coplanarity-only Robust Registration}
\label{sec:coponly}

At lines 526-529 of the main paper and in Table 1b, we provide an ablation study in which our method is compared to a variant (called ``Coplanarity only'') that uses only predicted matches of coplanar patches to constrain camera poses -- i.e., without keypoint matches.   In order to produce that one comparative result, we implemented an augmented version of our algorithm that includes a new method for selecting coplanar patch pairs in order to increase the chances of fully constraining the camera pose DoFs with the seletected patch pairs.   The following subsections describe that version of the algorithm.   Although it is not part of our method usually, we describe it in full here for the sake of reproducibility of the ``Coplanarity only'' comparison provided in the paper.

\subsection{Formulation}
\label{subsec:formulation}


\paragraph{Objective Function:}
The objective of coplanarity-only registration contains three terms,
including the \emph{coplanarity data term} (Equation (3) of the main paper),
the \emph{coplanarity regularization term} (Equation (4) of the main paper),
and a newly introduced \emph{frame regularization term} for regularizing the optimization
based on the assumption that the transformation between adjacent frames is small:
\begin{equation}
E({T},s) = E_{\text{data-cop}}({T},s)
+ E_{\text{reg-cop}}(s)
+ E_{\text{reg-frm}}({T})
\label{eq:robustopt}
\end{equation}

The frame regularization term $E_{\text{reg-frm}}$ makes sure the system is always solvable, by weakly constraining the transformations of adjacent frames to be as close as possible:
\begin{equation}
E_{\text{reg-frm}}(T) = \lambda\sum_{i \in \mathcal{F}}\sum_{\mathbf{v} \in \mathcal{V}_i}{||\mathbf{T}_i\mathbf{v}-\mathbf{T}_{i+1}\mathbf{v}||^2},
\label{eq:regfrm}
\end{equation}
where $\mathcal{V}_i$ is a sparse set of points sampled from frame $i$. $\lambda$~is set to $0.001$ by default.

When using coplanarity constraints only (without key-points),
our coplanarity-based alignment may become under-determined or unstable along some DoF,
when there are too few coplanar patch pairs that can be used to pin down that DoF.
In this case, we must be more willing to keep pairs constraining that DoF, to keep the system stable. To this end, we devise an anisotropic control variable, $\mu$, for patch pair pruning: If some DoF is detected to be unstable and enforcing $p_k$ and $q_k$ to be coplanar can constrain it, we set $\mu(\pi_k)$ to be large.
The alignment stability is estimated by analyzing the eigenvalues of the 6-DoF alignment error covariance matrix (gradient of the point-to-plane distances w.r.t.\ the six DoFs in $\mathbf{R}$ and $\mathbf{t}$) as in~\cite{gelfand03Stable}(See details in Section~\ref{subsec:stability}). 
Since the stability changes during the optimization, $\mu$ should be updated dynamically, and
we describe an optimization scheme with dynamically updated $\mu$ below.

\subsection{Optimization}
\label{subsec:optimization}

The optimization process is given in Algorithm~\ref{algo:opt}.
The core part is solving Equation (\ref{eq:robustopt}) via alternating optimization of
transformations and selection variables (the inner loop in Line~\ref{algol:inners}\textapprx\ref{algol:innere}).
The iterative process converges when the relative value change of each unknown is
less than $1 \times 10^{-6}$, which usually takes less than $20$ iterations.

A key step of the optimization is stability analysis and stability-based anisotropic pair pruning (Line~\ref{algol:stbs}\textapprx\ref{algol:stbe}). Since our coplanarity-based alignment is inherently orientation-based, it suffices to inspect the stability of the three translational DoFs. Given a frame $i$, we estimate its translational stability values, denoted by $\gamma^d_i$ ($d$ is one of the labels of X, Y, and Z-axis), based on the alignment of all frame pairs involving $i$ (see Section~\ref{subsec:stability} for details). One can check the stability of frame $i$ along DoF $d$ by examining whether the stability value $\gamma^d_i$ is greater than a threshold $\gamma_t$.

\IncMargin{0.5em}
\begin{algorithm}[tb]
\caption{Coplanarity-based Registration}
\label{algo:opt}
\small
\SetCommentSty{textsf}
\SetKwInOut{AlgoInput}{Input}
\SetKwInOut{AlgoOutput}{Output}
\SetKwFunction{Acquire}{Depth}
\SetKwFunction{Classify}{Classify}
\SetKwFunction{CompStab}{EstimateStability}
\SetKwFunction{Root}{RootNode}
\SetKwFunction{Child}{ChildNode}
\SetKwFunction{ChildShape}{ChildNodeShape}
\SetKwFunction{NBV}{RegressNBV}
\AlgoInput{
RGB-D frames $\mathcal{F}$ and co-planar patch pairs $\Pi = \cup_{(i,j)\in \mathcal{P}} \Pi_{ij}$;
$\gamma_t = 0.5$m.
}
\AlgoOutput{ Frame poses $T=\{(\mathbf{R}_{i}, \mathbf{t}_{i})\}$. }

$\mathbf{R}_{i}$ $\leftarrow$ $\mathbf{I}$, $\mathbf{t}_{i}$ $\leftarrow$ $\mathbf{0}$ \tcp*[r]{\scalebox{0.9}{Initialize transformations}}
$\mu^d_i$ $\leftarrow$ $0.1$m \tcp*[r]{\scalebox{0.9}{Initialize control variables}}
\Repeat {$\gamma_\emph{max} < \gamma_t$ \emph{or} max. \# of iterations reached} {
	\While { not converged } { \label{algol:inners}
    	Fix $s$, solve Equation (\ref{eq:robustopt}) for $T$ \;
        Fix $T$, solve Equation (\ref{eq:robustopt}) for $s$ \;
    } \label{algol:innere}
    $\{\gamma^d_i\} \leftarrow$ \CompStab($\Pi$, $s$) \; \label{algol:stbs}
    \ForEach (\tcp*[f]{\scalebox{0.9}{for each frame}}) {$i \in \mathcal{F}$} {
    	\ForEach (\tcp*[f]{\scalebox{0.9}{for each DoF}}) {$d \in \{\emph{X, Y, Z}\}$} {
    		\If {$\gamma^d_i > \gamma_t$} {
        		$\mu^d_i = \mu^d_i * 0.5$ \;
        	}
    	}
    }
    $\gamma_{\text{max}} \leftarrow \max_{i, d} \{ \gamma^d_i \}$ \; \label{algol:stbe}
}
\Return{$T$} \;
\end{algorithm}
\DecMargin{0.5em} 

Stability-based anisotropic pair pruning is achieved by dynamically setting the pruning parameter for a patch pair, $\mu(\pi)$ in the coplanarity regularization term (Equation (4) of the main paper). To this end, we set for each frame and each DoF an independent pruning parameter: $\mu^d_i$ ($i \in \mathcal{F}$ and $d=\text{X},\text{Y},\text{Z}$). They are uniformly set to a relatively large initial value ($0.1$m), and are decreased in each outer loop to gradually allow more pairs to be pruned. For some $\mu^d_i$, however, if its corresponding stability value $\gamma^d_i$ is lower than $\gamma_t$, it stops decreasing to avoid unstableness.
At any given time, the pruning parameter $\mu(\pi)$, with $\pi=(p,q)$, is set to:
$$
\mu(\pi) = \min\{\mu^{d(p)}_i, \mu^{d(q)}_j\},
$$
where $d(p)$ is the DoF closest to the normal of patch $p$. The whole process terminates when the stability of all DoFs becomes less than $\gamma_t$.

\begin{figure}[t!] \centering
    \begin{overpic}[width=0.9\linewidth]{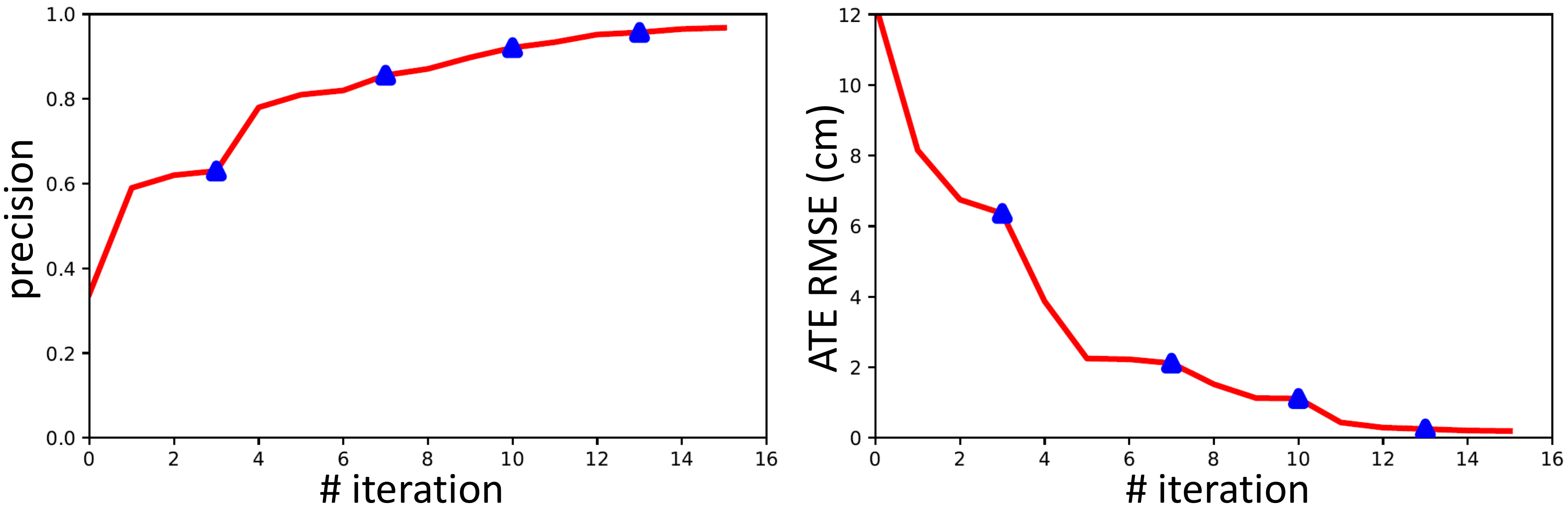}
    \end{overpic}
    \caption{The percentage of correct co-planar patch pairs increases and
    trajectory error (see the measure in Section 4 of the paper) decreases, as the iterative optimization proceeding. The blue marks indicate the outer loop.}
    \label{fig:iteration}
\end{figure}

To demonstrate the capability of our optimization to prune incorrect patch pairs,
we plot in Figure~\ref{fig:iteration} the ratio of correct coplanarity matches
at each iteration step for a ground-truth set.
We treat a pair $\pi$ as being kept if its selection variable $s(\pi)>0.5$ and discarded otherwise.
With more and more incorrect pairs pruned, the ratio increases while the registration error (measured by absolute
camera trajectory error (ATE); see Section 4 of the paper) decreases.

\subsection{Stability Analysis}
\label{subsec:stability}
The stability analysis of coplanar alignment is inspired by the work of Gelfand et al.~\cite{gelfand03Stable} on geometrically stable sampling for point-to-plane ICP.
Consider the point-to-plane alignment problem found in the data term of our coplanarity-based registration (see Equation (3) in the main paper).
Let us assume we have a collection of points $\mathbf{v}_p \in \mathcal{V}_p$ sampled from patch $p$,
and a plane $\phi_q = (\mathbf{p}_q, \mathbf{n}_q)$ defined by patch $q$.
We want to determine the optimal rotation and translation to be applied to the point set $\mathcal{V}_p$,
to bring them into coplanar alignment with the plane $\phi_q$.
In our formulation, source and target patches ($p$ and $q$) are also exchanged to compute alignment error bilaterally
(see Line 436 in paper). Below we use only patch $p$ as the source for simplicity of presentation.

We want to minimize the alignment error
\begin{equation}
\mathcal{E} = \sum_{\mathbf{v}_p \in \mathcal{V}_p}{\bigl[(\mathbf{R}\mathbf{v}_p+\mathbf{t}-\mathbf{p}_q) \cdot \mathbf{n}_q\bigr]^2},
\label{eq:error}
\end{equation}
with respect to the rotation $\mathbf{R}$ and translation $\mathbf{t}$.

The rotation is nonlinear, but can be linearized by assuming that incremental rotations will be small:
\begin{equation}
\mathbf{R} \approx \begin{pmatrix}
                     1 & -r_z & r_y \\
                     r_z & 1 & -r_x \\
                     -r_y & r_x & 1 \\
                   \end{pmatrix},
\label{eq:rotlinear}
\end{equation}
for rotations $r_x$, $r_y$, and $r_z$ around the X, Y, and Z axes, respectively.
This is equivalent to treating the transformation of $\mathbf{v}_p \in \mathcal{V}_p$ as a
displacement by a vector $[\mathbf{r}\times \mathbf{v}_p + \mathbf{t}]$, where
$\mathbf{r}=(r_x, r_y, r_z)$. Substituting this into Equation (\ref{eq:error}),
we therefore aim to find a 6-vector $[\mathbf{r}^T, \mathbf{t}^T]$ that minimizes:
\begin{equation}
\mathcal{E} = \sum_{\mathbf{v}_p \in \mathcal{V}_p}{(\mathbf{v}_p-\mathbf{p}_q) \cdot \mathbf{n}_q + \mathbf{r} \cdot (\mathbf{v}_p \times \mathbf{n}_q) + \mathbf{t} \cdot \mathbf{n}_q}.
\label{eq:linearerror}
\end{equation}

We solve for the aligning transformation by taking partial derivatives of Equation (\ref{eq:linearerror}) with respect
to the transformation parameters in $\mathbf{r}$ and $\mathbf{t}$.
This results in a linear system $C\mathbf{x}=\mathbf{b}$ where $\mathbf{x}=[\mathbf{r}^T, \mathbf{t}^T]$
and $\mathbf{b}$ is the residual vector. $C$ is a $6\times 6$ ``covariance matrix'' of the rotational
and translational components, accumulated from the sample points:
\begin{equation*}
C = \begin{bmatrix}
      \mathbf{v}_p^1 \times \mathbf{n}_q & \cdots & \mathbf{v}_p^k \times \mathbf{n}_q \\
      \mathbf{n}_q & \cdots & \mathbf{n}_q \\
    \end{bmatrix}
    \begin{bmatrix}
      (\mathbf{v}_p^1 \times \mathbf{n}_q)^T & \mathbf{n}_q \\
      \vdots & \vdots \\
      (\mathbf{v}_p^k \times \mathbf{n}_q)^T & \mathbf{n}_q \\
    \end{bmatrix}.
\end{equation*}
This covariance matrix encodes the increase in the alignment error due to the movement of
the transformation parameters from their optimum:
\begin{equation}
\Delta \mathcal{E} = 2
    \begin{bmatrix}
        \Delta \mathbf{r}^T & \Delta \mathbf{t}^T \\
    \end{bmatrix}
    C
    \begin{bmatrix}
        \Delta \mathbf{r} \\
        \Delta \mathbf{t} \\
    \end{bmatrix}.
\label{eq:derive}
\end{equation}

The larger this increase, the greater the stability of the alignment, since the error landscape will have a deep,
well-defined minimum. On the other hand, if there are incremental transformations that cause only
a small increase in alignment error, it means the alignment is relatively unstable along that degree of freedom.
The analysis of stability can thus be conducted by finding the eigenvalues of matrix $C$. Any small
eigenvalues indicate a low-confidence alignment. In our paper, we analyze translational stabilities
based on the eigenvalues corresponding to the three translations, $\gamma^d$ ($d=\text{X, Y, Z}$).

\end{document}